%% file: main.tex
\pdfoutput=1

\documentclass[11pt]{article}

\usepackage[final]{acl}

\usepackage{times}
\usepackage{latexsym}
\usepackage{graphicx}
\usepackage[T1]{fontenc}

\usepackage[utf8]{inputenc}

\usepackage{microtype}

\input{00_definition}

\title{Rethinking Round-Trip Translation for Machine Translation Evaluation}

\author{Terry Yue Zhuo$^1$ \and Qiongkai Xu$^2$\thanks{*Corresponding author} \and Xuanli He$^3$ \and Trevor Cohn$^2$ \\
            $^1$ Monash University, Clayton, VIC, Australia \\
        $^2$ The University of Melbourne, Carlton, VIC, Australia \\
        $^3$ University College London, London, United Kingdom\\
        \texttt{terry.zhuo@monash.edu} \\ \texttt{\{qiongkai.xu,trevor.cohn\}@unimelb.edu.au}\\
        \texttt{zodiac.he@gmail.com}
        }

\begin{document}
\maketitle

\input{01_abstract}
\input{02_introduction}

\input{03_related_work}
\input{04_smt_rrt}

\input{07_1_exp_pred_FT}

\input{07_3_exp_improve_QE}

\input{07_2_exp_cross_system}
\input{10_conclusion}
\bibliography{custom}
\bibliographystyle{acl_natbib}

\newpage
\clearpage
\appendix

\input{08_appendix}

\end{document}

%% file: 00_definition.tex
\usepackage{soul}
\usepackage{stmaryrd}
\usepackage{amsfonts}
\usepackage{amssymb}
\usepackage{amsmath,mathtools}
\usepackage{bbm}
\usepackage{multirow,tabularx}
\usepackage{xspace}
\usepackage{todonotes}
\usepackage{booktabs}
\usepackage{subcaption}
\usepackage{makecell, tabularx}
\usepackage{arydshln}
\usepackage{pifont}
\usepackage{diagbox}

\definecolor{deemph}{gray}{0.6}

\definecolor{green}{HTML}{39b54a}  
\definecolor{greenn}{HTML}{008000}  
\definecolor{blue}{HTML}{1F51FF}  %

\newcommand{\cmark}{\texttt{X-Check}\xspace}
\newcommand{\xmark}{\texttt{Sing-Check}\xspace}

\newcommand{\dataset}{\texttt{FLORES-AE33}\xspace}
\newcommand{\flores}{\texttt{FLORES-101}\xspace}
\newcommand{\wmt}{\texttt{WMT}\xspace}
\newcommand{\newscommentary}{\texttt{News-Commentary v8}\xspace}

\newcommand{\wmtnews}{\texttt{WMT2020-News}\xspace}
\newcommand{\wmtbio}{\texttt{WMT2020-Bio}\xspace}

\newcommand{\moses}{\texttt{Moses}\xspace}

\newcommand{\typeI}{\textbf{Type~I}\xspace}
\newcommand{\typeII}{\textbf{Type~II}\xspace}
\newcommand{\typeIII}{\textbf{Type~III}\xspace}

\newcommand{\transscore}{\textsc{FT-Score}\xspace}
\newcommand{\selfscore}{\textsc{RT-Score}\xspace}
\newcommand{\transscores}{\textsc{FT-Score}s\xspace}
\newcommand{\selfscores}{\textsc{RT-Score}s\xspace}

\newcommand{\rtt}{\text{RTT}\xspace}
\newcommand{\ft}{\text{FT}\xspace}

\newcommand{\bt}{\text{BT}\xspace}

\newcommand{\MaxCount}{\textsc{Max-4 Count}\xspace}
\newcommand{\RefLen}{\textsc{Ref Length}\xspace}

\newcommand{\mbart}{\textsc{mBART50-m2m}\xspace}
\newcommand{\mmbase}{\textsc{M2M-100-base}\xspace}
\newcommand{\mmlarge}{\textsc{M2M-100-large}\xspace}
\newcommand{\googletrans}{\textsc{Google-Trans}\xspace}
\newcommand{\opus}{\textsc{Opus-MT}\xspace}
\newcommand{\marian}{\textsc{Marian-NMT}\xspace}
\newcommand{\easynmt}{\textsc{EasyNMT}\xspace}
\newcommand{\bleutool}{\textsc{sacreBLEU-Toolkit}\xspace}
\newcommand{\berttool}{\textsc{BERTScore}\xspace}

\newcommand{\sacrebleu}{\texttt{BLEU}\xspace}
\newcommand{\spbleu}{\texttt{spBLEU}\xspace}
\newcommand{\chrf}{\texttt{chrF}\xspace}
\newcommand{\bertscore}{\texttt{BERTScore}\xspace}

\newcommand{\cometda}{\texttt{COMET-QE-DA}\xspace}

\newcommand{\xcheck}{\texttt{X-Check}\xspace}

\newcommand{\seen}{\ensuremath{%
  \mathchoice{\includegraphics[height=2ex]{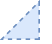}}
    {\includegraphics[height=2ex]{figures/seen.pdf}}
    {\includegraphics[height=1.5ex]{figures/seen.pdf}}
    {\includegraphics[height=1ex]{figures/seen.pdf}}
}}

\newcommand{\unseen}{\ensuremath{%
  \mathchoice{\includegraphics[height=2ex]{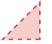}}
    {\includegraphics[height=2ex]{figures/unseen.pdf}}
    {\includegraphics[height=1.5ex]{figures/unseen.pdf}}
    {\includegraphics[height=1ex]{figures/unseen.pdf}}
}}

%% file: 01_abstract.tex
\begin{abstract}

Automatic evaluation on low-resource language translation suffers from a deficiency of parallel corpora. Round-trip translation could be served as a clever and straightforward technique to alleviate the requirement of the parallel evaluation corpus. However, there was an observation of obscure correlations between the evaluation scores by forward and round-trip translations in the era of statistical machine translation (SMT).
In this paper, we report the surprising finding that round-trip translation can be used for automatic evaluation without the references. Firstly, our revisit on the round-trip translation in SMT evaluation unveils that its long-standing misunderstanding is essentially caused by copying mechanism. After removing copying mechanism in SMT, round-trip translation scores can appropriately reflect the forward translation performance. Then, we demonstrate the rectification is overdue as round-trip translation could benefit multiple machine translation evaluation tasks. To be more specific, round-trip translation could be used i) to predict corresponding forward translation scores; ii) to improve the performance of the recently advanced quality estimation model; and iii) to identify adversarial competitors in shared tasks via cross-system verification.     

\end{abstract}

%% file: 02_introduction.tex
\section{Introduction}

Thanks to the recent progress of neural machine translation (NMT) and large-scale multilingual corpora, 
machine translation (MT) systems have achieved remarkable performances on high- to medium-resource languages \cite{fan2021beyond, pan2021contrastive,flores101}. However, the development of MT technology on low-resource language pairs still suffers from insufficient data for training and evaluation~\cite{aji2022one,siddhant2022towards}. 
Recent advances in  multilingual pre-trained language model explore the methods trained on monolingual data, using data augmentation and denoising auto-encoding method~\cite{xia2019generalized,liu2020multilingual}. However, high-quality parallel corpora are still required for evaluating translation quality. Such requirement is especially resource-consuming when working on \textit{i)} hundreds of underrepresented low-resource languages~\citep{bird2012machine,joshi2019unsung,aji2022one} and \textit{ii)} translations for specific domains~\citep{li2020metamt,muller2019domain}.
\begin{figure}
    \centering
    \includegraphics{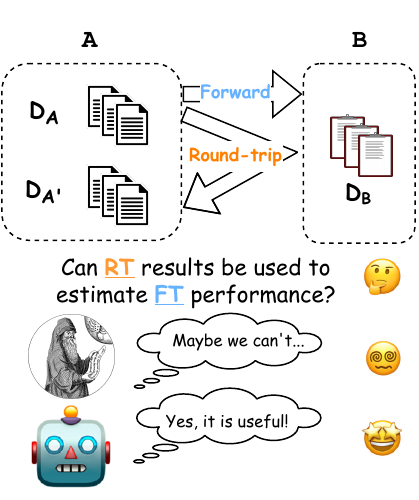}
    \caption{Given the corpus \textit{\textbf{$D_A$}} in Language \textbf{$A$}, we are able to acquire the round-trip translation (RT) results \textit{\textbf{$D_A^{'}$}} and forward translation (FT) results \textit{\textbf{$D_B$}} via machine translation. A question was raised by machine translation community two decades ago, ``Can \emph{RT} results be used to estimate \emph{FT} performance?''. The answers to the question appeared to be \emph{controversial}, where most researchers tend to be against the use of round-trip translation due to the obscure correlations between them. Our work gives a clear and positive answer to the usefulness of RT, based on extensive experiments.}
    \label{fig:teaser}
    \vspace{-3mm}
\end{figure}
In order to mitigate the deficiency of parallel corpora, conducting Round-trip Translation (\rtt) could be a promising method for training data augmentation and evaluation solely on the monolingual corpus. As illustrated in Figure~\ref{fig:teaser}, \rtt entails two components, one forward translation (\ft), and the other backward translation (\bt). \ft translates a given sentence in source language $A$ to a sentence in target language $B$, then the output sentence from \ft is translated back to language $A$ via a back translation system. 
However, the existing literature demonstrates that the automatic evaluation score on \rtt (\selfscore) unfortunately fails to reflect the score of \ft quality (\transscore) on statistical machine translation (SMT) and rule-based machine translation (RMT) systems~\cite{huang1990machine,koehn2005europarl,somers2005round,zaanen2006unsupervised}. This understanding impedes the usage of \rtt for MT evaluation on monolingual data, until some recent empirical discovery of \rtt could be helpful for quality estimation (QE) using sentence embeddings~\cite{moon-etal-2020-revisiting, crone2021quality}. More recently, a study~\cite{agrawal2022quality} has shown that \rtt-based  QE without sentence embeddings can also complement state-of-the-art QE systems.
In this work, we revisit the dispute on the usefulness of \selfscore in the era of SMT versus NMT. The main reason is due to the fact that SMT (and RMT) usually incorporate implicitly reversible rules in their translation. For example, copying unrecognized tokens forward to target languages is sometimes penalized by \ft evaluation while it is usually awarded by \rtt evaluation. Extensive experiments are conducted to demonstrate the effect of copying mechanism on SMT. Later, we illustrate strong correlations between \transscores and \selfscores on various MT systems, including NMT and SMT without the copying mechanism. 

The finding sets the basis of using \selfscore for MT evaluation. Three application scenarios in MT evaluation have been investigated to show the effectiveness of \selfscore. 
Firstly, \selfscores can be used to predict \transscores via training a simple but effective linear regression model on several hundred language pairs. The prediction performance is robust in evaluating transferred MT systems and unseen language pairs including low-resource languages. 
Then, a cross-system check (\xcheck) mechanism is introduced to \rtt evaluation for real-world MT shared tasks. By leveraging the estimation from multiple translation systems, \xcheck manages to identify those adversarial competitors, which relies heavily on the copy strategy. Finally, \selfscores are proved effective in improving the performance of a recently advanced quality estimation model. 



%% file: 03_related_work.tex
\section{Related Work}
\paragraph{Reference-based Machine Translation Evaluation Metric.}
Designing high-quality automatic evaluation metrics for evaluating translation quality is one of the fundamental challenges in MT research. Most of the existing metrics largely rely on parallel corpora to provide aligned texts as references~\cite{papineni2002bleu,lin2004rouge}. The performance of the translation is estimated by comparing the system outputs against ground-truth references. 
A classic school of reference-based evaluation is based on string match method, which calculates matched ratio of strings or lexicon, such as BLEU~\cite{papineni2002bleu, post-2018-call}, ChrF~\cite{popovic2015chrf} and TER~\cite{snover2006study}. 
In addition, recent metrics utilize the semantic representations of texts to estimate their relevance, given pre-trained language models, such as BERTScore~\cite{zhang2019bertscore} and BLEURT~\cite{sellam2020bleurt}. These methods are demonstrated to be more correlated to human evaluation~\cite{kocmi2021ship} than string-based metrics. Some other reference-based evaluation metrics require supervised training to work well \cite{mathur2019putting, rei2020comet} on contextual
word embeddings. While these automatic evaluation metrics are widely applied in MT evaluation, they are generally not applicable to low-resource language translation or new translation domains~\cite{mathur2020tangled}. Our work demonstrates that reference-free MT metrics (\selfscore) could be used to estimate traditional reference-based automatic metrics. 

\paragraph{Reference-free Quality Estimation.}
In recent years, there has been a surge of interest in the task of directly predicting the human judgment, namely quality estimation (QE), without the access to parallel reference translations in the run-time~\cite{specia2010machine, specia2013quest, bojar2014findings, zhao2020limitations}. Recent focus on QE is mainly based on human evaluation approaches, direct assessment (DA) and post-editing (PE), where researchers intend to train models on data via human judgement features to estimate MT quality. 
Among these recent QE metrics, learning based models, YiSi-2~\cite{lo2019yisi}, COMET-QE-MQM~\cite{rei2021references}, to name a few, demonstrate their effectiveness on \wmt shared tasks. Our work shows \selfscore promote a recently advanced QE model.

%% file: 04_smt_rrt.tex
\begin{figure*}[th]
    \centering
    \includegraphics[width=0.95\linewidth]{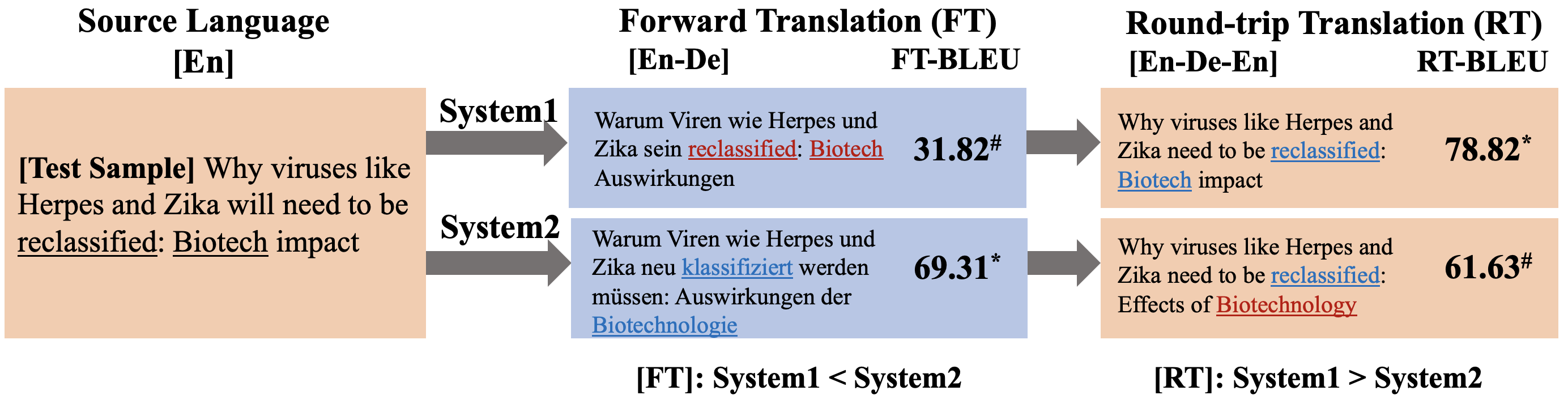}
    \caption{The comparison of the forward translation (FT) and round-trip translation (RT) performance of two translation systems, System 1 and System 2 are based on Statistical Machine Translation (SMT) and Neural Machine Translation (NMT), respectively. The conflict conclusions by FT Scores (System 1 $<$ System 2) and RT Scores (System 1 $>$ System 2) are attributed to the translation of the underlined words, `reclassified' and `Biotech'.} 
    \label{fig:rt_vs_ft}
    \vspace{-3mm}
\end{figure*}

\section{Revisiting Round-trip Translation}
\subsection{Evaluation on Round-trip Translation}
Given machine translation systems, $\mathcal T_{A\rightarrow B}$ and $\mathcal T_{B\rightarrow A}$, between two languages ($L_A$ and $L_B$), and a monolingual corpus $\mathcal{D}_A=\{a_i\}_{i=1}^{N}$, \ft transforms $a_i$ to $b'_i=\mathcal T_{A\rightarrow B}(a_i)$ and \bt translates it back to $A$, $a'_i=\mathcal T_{B\rightarrow A}(\mathcal T_{A\rightarrow B}(a_i))$. \ft and \bt constitute a round-trip translation (\rtt).

The evaluation scores on round-trip translation (\selfscore) with regard to an automatic evaluation metric $\mathcal M$ is 
\begin{equation}
\begin{small}
    \text{\textsc{\selfscore}}\text{\textsc{}}_{A\circlearrowright B}^{\mathcal{M}} = 
    \frac{1}{N}\sum_{i=1}^{N}\mathcal{M} (\mathcal T_{B\rightarrow A}(\mathcal T_{A\rightarrow B}(a_i)), a_i)
\end{small}
\end{equation}
where \sacrebleu, \spbleu, \chrf and \bertscore are target metrics $\mathcal M$ in our discussion.

On the other hand, traditional MT evaluation on parallel corpus is
\begin{equation}
\begin{small}
    \text{\transscore}_{A\rightarrow B}^{\mathcal{M}}=
    \frac{1}{N}\sum_{i=1}^{N}\mathcal{M} (\mathcal T_{A\rightarrow B}(a_i), b_i)
\end{small}
\end{equation}
given a (virtual) parallel corpus $\mathcal{D}_{A||B}=\{(a_i, b_i)\}_{i=1}^{N}$. 
The main research question is whether \transscores are correlated to therefore could be predicted by \selfscores.

\subsection{\rtt Evaluation on Statistical Machine Translation}
\label{sec:smt_exp}
The previous analysis on the automatic evaluation scores from \rtt and \ft shows that they are negatively correlated. Such a long-established understanding started from the era of RMT~\citep{huang1990machine} and lasted through SMT~\cite{koehn2005europarl,somers2005round} and prevented the usage of \rtt to MT evaluation. We argue that the negative observations are probably due to the selected SMT models involving some reversible transformation rules, e.g., copying unrecognized tokens in translation. As an example illustrated in Figure~\ref{fig:rt_vs_ft}, the MT System 1 works worse than its competing System 2, as System 1 fails to translate `reclassified' and `Biotech'. Instead, it decides to copy the words in source language (En) directly to the target outputs. During \bt, System 1 manages to perfectly translate them back without any difficulty. For System 2, although translating `Biotechnologie' (De) to `Biotechnology' (En) is adequate, it is not appreciated by the original reference in this case. Consequently, the rankings of these two MT systems are flipped according to their \ft and \rtt scores. 
Previous error analysis study on SMT~\cite{vilar2006error} also mentioned that the unknown word copy strategy is one of the major causes resulting in the translation errors. We therefore argue that the reversible transformation like word copy could have introduced significant bias to the previous experiments on SMT (and RMT). Then, we conduct experiments to replicate the negative conclusion. Interestingly, removing the copying mechanism can almost perfectly resolve the negation in our experiments.

\subsection{Experiments and Analysis}
\label{sec:smt_vs_nmt_exp}
We compare \rtt and \ft on SMT following the protocol by \citet{somers2005round,koehn2005europarl}. 
\moses~\cite{koehn2009moses} is utilized to train phrase-based MT systems~\cite{koehn2003statistical}, which were popular in the SMT era.\footnote{We follow the baseline setup in the Moses' tutorial in \url{http://www2.statmt.org/moses/?n=Moses.Baseline}.} We train SMT systems on \newscommentary \cite{TIEDEMANN12.463}, as suggested by \wmt organizers~\cite{koehn2006proceedings}. We test our systems on six language pairs (de-en, en-de, cs-en, en-cs, fr-en and en-fr) in the competition track of \wmt Shared Tasks~\cite{barrault-etal-2020-findings}. \selfscores and \transscores are calculated based on \sacrebleu in this section. Then, we use Kendall's $\tau$ and Pearson's $r$ to verify the correlation of \selfscores and \transscores~\cite{kendall1938new, benesty2009pearson}. 
We provide more detailed settings in Appendix~\ref{appendix:mts_detail}.

During translation inference, we consider two settings for comparison, one drops the unknown words and the other one copies these tokens to the outputs. Hence, we end up having two groups of six outputs from various SMT systems.  

\input{tables/smt_copy_word}


In Table~\ref{tab:smt_copy_word}, we examine the relevance between \selfscores and \transscores on six SMT systems. The performance is measured by Kendall's $\tau$ and Pearson's $r$. The correlation is essentially decided by the copying mechanism. Specifically, their correlation turns to be much stronger for those systems not allowed copying, compared to the systems with default word copy. 

Now, we discuss the rationality of using \rtt evaluation for NMT systems, by comparing the reliance of copying mechanism in NMT and SMT. For NMT, we choose \mbart~\cite{tang2020multilingual}, which covers 50 languages of cross-lingual translation. Exactly matched words in outputs from the input words are considered copying, although the system may not intrinsically intend to copy them. In Table~\ref{tab:word_copy}, we observe that copying frequency is about two times in SMT than in NMT. Although NMT systems may copy some words during translation, most of them are unavoidable, e.g., we observe that most of these copies are proper nouns whose translation are actually the same words in target language. In contrast, the copied words in SMT are more diverse and many of them could be common nouns. 
\input{tables/word_copy_smt_nmt}

%% file: tables/smt_copy_word.tex
\begin{table}[h]
\small
\centering
\begin{tabular}{l|cc|cc}
\hline
\multirow{2}{*}{Lang. Pair}  & \multicolumn{2}{c|}{\textbf{K.} $\boldsymbol \tau$ $\uparrow$} & \multicolumn{2}{c}{\textbf{P.} $\boldsymbol r$ $\uparrow$}  \\\cline{2-5} 
& w/ cp & w/o cp & w/ cp & w/o cp  \\ \hline
de-en   & -0.11& 0.20 &-0.90 & 1.00\\\hline
en-de & -0.40& 0.60 & -1.00 & 1.00\\\hline
cs-en  & -0.20& 0.30 & -0.99 & 0.99\\\hline
en-cs  & -0.40 & 0.60 & -0.90 & 0.99\\\hline
fr-en  & 0.20 & 0.60 & -1.00 & 1.00\\\hline
en-fr  & -0.40 & 1.00 & -0.90 & 0.99\\\hline

\end{tabular}
\caption{Comparison between \selfscore and \transscore on two groups of systems with copying (w/ cp) and without copying (w/o cp) unknown words using Kendall's $\tau$ and Pearson's $r$ on six language pairs.} 
\label{tab:smt_copy_word}
\vspace{-2mm}
\end{table}

%% file: tables/word_copy_smt_nmt.tex
\begin{table}[!ht]
\small
\centering

\begin{tabular}{c|cc}
\hline
   \multirow{2}{*}{Lang. Pair}  & \multicolumn{2}{c}{Avg. Copy (\%)} \\\cline{2-3}
& SMT & NMT \\\hline
   de-en  & 17.39 & 9.28\\
   en-de  & 21.47 &9.54\\\hline
\end{tabular}
\caption{Comparison of word copy frequency between SMT and NMT on two language pairs. We calculate average percentage of copy (Avg. Copy) per sentence. The details of selected \moses system are reported in Appendix~\ref{appendix:mts_detail}.}
\label{tab:word_copy}
\vspace{-5mm}
\end{table}


%% file: 07_1_exp_pred_FT.tex

\section{Predicting \transscore using \selfscore}
In this section, we validate whether \transscores could be predicted by \selfscores. Then, we examine the robustness of the predictor on unseen language pairs and transferred MT models.

\subsection{Regression on \selfscore} 

Here, we construct a linear regressor $f$ to predict \transscores of a target translation metric $\mathcal M$ by corresponding \selfscores,
\begin{align}
    \nonumber \text{\transscore}_{A\rightarrow B}^{\mathcal{M}} \approx 
     f_{\mathcal{M}}(&\text{\selfscore}_{A\circlearrowright B}^{\mathcal{M}^*},\\
     &\text{\selfscore}_{B\circlearrowright A}^{\mathcal{M}^*}).
\end{align}
$\mathcal{M}^*$ indicates that multiple metrics could be used to construct the input features. We utilize \selfscore from both sides of a language pair as our primary setting, as using more features usually provides better prediction performance~\cite{xia2020predicting}.
We introduce a linear regressor for predicting \transscore,
\begin{equation}
    f_{\mathcal{M}}(\mathbf{S}) = \mathbf{W}_1 \cdot \mathbf{S}^{\mathcal{M}^{*}}_{A \circlearrowright B} + \mathbf{W}_2 \cdot \mathbf{S}^{\mathcal{M}^{*}}_{B \circlearrowright A} + \beta
\end{equation}
where $\mathbf{S}^{\mathcal{M}^{*}}_{A \circlearrowright B}$ and $\mathbf{S}^{\mathcal{M}^{*}}_{B \circlearrowright A}$ are \selfscore features used as inputs of the regressor\footnote{We use $\mathcal{M}^{*}=\mathcal{M}$ as our primary setting, as it is the most straightforward and effective method to construct features. In addition, we discuss the possibility to improve the regressor by involving more features, in Appendix~\ref{appendix:more_features}.}. $\mathbf{W}_1$, $\mathbf{W}_2$ and $\beta$ are the parameters of the prediction model optimized by supervised training.~\footnote{Implementation details can be found in Appendix~\ref{appendix:implement}.}

In addition, when organizing a new shared task, say \wmt, collecting a parallel corpus in low-resource language 
could be challenging and 
resource-intensive. Hence, we investigate another setting that utilizes merely the monolingual corpora in language $A$ or $B$ to predict \transscore,
\begin{align} \nonumber
    \text{\transscore}_{A\rightarrow B}^{\mathcal{M}} \approx  f'_{\mathcal{M}}(\text{\selfscore}_{A\circlearrowright B}^{\mathcal{M}^*}),\\
    \text{\transscore}_{A\rightarrow B}^{\mathcal{M}} \approx  f'_{\mathcal{M}}(\text{\selfscore}_{B\circlearrowright A}^{\mathcal{M}^*}).
    \label{eq:single_direct}
\end{align}

We will compare and discuss this setting in our experiments on \wmt.

\label{sec:application_1}
\subsection{Experimental Setup}
\subsubsection{Datasets}
We conduct experiments on the large-scale multilingual benchmark, \flores, and \wmt machine translation shared tasks. \dataset is for training and testing on  languages and transferred MT systems. \wmt is for testing on real-world shared tasks in new domains.

\begin{figure}
    \centering
     \includegraphics[width=0.7\linewidth]{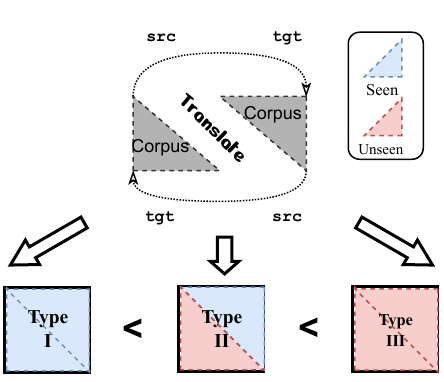}
    \caption{The 33 languages in \dataset are separated into two categories, \texttt{Seen}\xspace (\seen) includes the languages used in both \emph{training} and \emph{testing}, and \texttt{Unseen}\xspace (\unseen) is composed of the languages only used in \emph{testing}. \seen contains 7 High-resource (H.), 7 Medium-resource (M.) and 6 Low-resource (L.) languages, while \unseen involves 9 Medium-resource (M.) and 4 Low-resource (L.) languages. These two sets are used to construct three types of language pairs for test. \typeI and \typeIII target on translation among \seen and \unseen language pairs, respectively. \typeII targets on translation between \seen and \unseen. The test setting with more \unseen is usually more challenging, i.e., \typeI $<$ \typeII $<$ \typeIII.}
    \label{fig:flores_ae33}
    \vspace{-3mm}
\end{figure}

\paragraph{\dataset.}

We extract \dataset, which contains parallel data among 33 languages, covering 1,056 ($33\times32$) language pairs, from a curated subset of \flores~\citep{flores101}. We select these languages based on two criteria: \textit{i)} We rank languages given the scale of their bi-text corpora; \textit{ii)} We prioritize the languages covered by \wmtnews and \wmtbio.
As a result, \dataset includes 7 high-resource languages, 16 medium-resource languages and 10 low-resource languages. We show the construction pipeline in Figure~\ref{fig:flores_ae33}, with more details in Appendix~\ref{appendix:dataset_detail}.

\paragraph{\wmt.} 
We collect corpora from the translation track to evaluate multiple MT systems on the same test sets. We consider their ranking based on \transscore with metric $\mathcal M$ as the ground truth. %
We choose the competition tracks in \wmt 2020 Translation Shared Tasks~\cite{barrault-etal-2020-findings}, namely news track \wmtnews and biomedical track \wmtbio. We consider \textit{news} and \textit{bio} as new domains, compared to our training data \flores whose contents are mostly from Wikipedia.


\subsubsection{Neural Machine Translation Systems}
We experiment with five MT systems which support most of the languages appearing in \dataset and \wmt. Except \mbart, we adopt \mmbase and \mmlarge~\cite{fan2021beyond}, which are proposed to conduct many-to-many MT without explicit pivot languages, supporting 100 languages. \googletrans~\cite{wu2016google, bapna2022building}\footnote{We queried \googletrans API in August, 2022.} is a commercial translation API, which was considered as a baseline translation systems in many previous competitions~\cite{barrault-etal-2020-findings}. Meanwhile, we also include a family of bilingual MT models, \opus~\cite{tiedemann2020opus}, sharing the same model architecture \marian~\cite{junczys2018marian}. We provide more details about these MT systems in Appendix~\ref{appendix:mts_detail}. 

\subsubsection{Automatic MT Evaluation Metrics}
We consider \sacrebleu, \spbleu~\cite{goyal2022flores}, \chrf~\cite{popovic2015chrf} and \bertscore~\cite{zhang2019bertscore} as the primary automatic evaluation metrics~\cite{freitag2020bleu}. All these metrics will be used and tested for both input features and target \transscore. The first two metrics are differentiated by their tokenizers, where \sacrebleu uses Moses~\cite{koehn2010moses} and \spbleu uses SentencePiece~\cite{kudo2018sentencepiece}. Both evaluation metrics were officially used in WMT21 Large-Scale Multilingual Machine Translation Shared Task~\cite{wenzek2021findings}. While \sacrebleu works for most language tokenizations, \spbleu shows superior effectiveness on various language tokenizations, especially the performance on low-resource languages~\cite{flores101}. More details of these metrics are described in Appendix~\ref{appendix:evaluation_metrics}\


\subsection{Experiments and Analysis}

\label{sec:experiments}


Following our discussion in the last section on SMT, we conduct similar experiments using our new multilingual NMT systems on \typeI test set of \dataset. We observe highly positives correlation between \transscores and \selfscores, measured by Pearson's $r$~\cite{benesty2009pearson}. Please refer to Appendix~\ref{appendix:rq_1} for more details. Then, we train regressors on \selfscores and conduct experiments to examine their performance on various challenging settings.


\input{tables/exp_system_transfer}

\input{tables/exp_language_transfer}
\begin{table}[]
    \centering
\resizebox{1.0\linewidth}{!}{
    \begin{tabular}{c|c|ccc | ccc} 
    \hline
     \multicolumn{2}{c|}{} & \multicolumn{3}{c|}{\textbf{MAE} $\downarrow$} & \multicolumn{3}{c}{\textbf{P.} $\boldsymbol r$ $\uparrow$}  \\\cline{3-8}
    
    \multicolumn{2}{c|}{} & H. & M. & L. & H. & M. & L. \\\hline
    \multirow{3}{*}{\mbart}
    & H. & 3.17 &2.90&2.70 &0.94 & 0.94 &0.77
\\
    & M. & 1.51 &1.37& 1.77&0.97 &0.85& 0.22\\
    & L & 1.22 &1.27& 1.16&0.97 &0.87&0.78\\\hline
    \multirow{3}{*}{\mmbase} & H. & 8.72 &5.41&3.50 &0.51 & 0.80 &0.67
\\
    & M. & 4.86 &4.01& 2.93&0.86 &0.90& 0.69\\
    & L & 1.70 &1.67& 1.24&0.98 &0.97&0.80\\\hline
    \end{tabular}}
    \caption{The results of predicted \transscores of \mbart and \mmbase on nine sets of language pairs, categorized by different scale of the resources, High (H.), Medium (M.) and Low (L.). The three categories in rows are source languages, and the ones in columns are target languages. We report Mean Absolute Error (MAE) and Pearson's $r$.}
    \label{tab:lang_resource_mbart}
    \vspace{-3mm}
\end{table}
\subsubsection{Transferability of Regressors}
\label{rq2}
We firstly investigate the transferability of our regressors from two different aspects, transferred MT systems and unseen language pairs. We also evaluate the regressor on different scales of language resources.

\paragraph{Settings.} We train our regressors on \typeI train set based on the translation scores from \mbart. In order to assess system transferability, we test three models
on \typeI test set. In terms of the language transferability, we consider \transscores of \mbart (a seen MT system in training) and \mmbase (an unseen MT system in training) on \typeII and \typeIII in \dataset. We further evaluate the transferability of our regressor on language resources in \typeI test set, with two MT systems, \mbart and \mmbase.


\input{tables/wmt20_news_results}
\paragraph{Discussion.} In Table~\ref{tab:system_transfer}, we present the performance of the regressor across various translation systems and evaluation metrics. We first analyze the results on \mbart, which is seen in training. The absolute errors between predicted scores and ground-truth \transscores are relatively small with regard to MAE and RMSE. Meanwhile, the correlation between prediction and ground truth is strong, with all Pearson's $r$ above or equal to 0.88. This indicates that the rankings of predicted scores are rational.
The results of \mmbase and \googletrans demonstrate the performance of predictors on \textit{unseen} systems. 
Although the overall errors are higher than those of \mbart without system transfer, Pearson's $r$ scores are at the competitive level, indicating a similar ranking capability on unseen systems.  
Meanwhile, our model obtains adequate language transferability results, as demonstrated in Table~\ref{tab:language_transfer}.

In Table~\ref{tab:lang_resource_mbart}, we provide detailed performance of our regressor on language pairs of different resources categories on \dataset, with \selfscores of \mbart and \mmbase respectively. Specifically, we split the three categories based on Table~\ref{tab:flores_ae33}, which are high, medium and low. 
The evaluated regressor is the same as the one tested in Sections \ref{rq2} and \ref{rq3}. The results of two tables show that our regressor is able to predict \transscores with small errors, and reflect the relative orders among \transscores, with high transferability across language pairs and MT systems.


\subsubsection{Predicting \transscores on WMT}
\label{rq3}
With the basis of high transferabilities of the regressors, we conduct experiments on \wmt shared tasks, namely \wmtnews, which includes 10 language pairs. In this experiment, we study on \spbleu metric scores.
\paragraph{Settings.} 
We have involved five MT systems
\footnote{We have contacted the competitors to \wmtnews. However, we have not received enough valid MT systems to increase the number of competitors. We will show the robustness of our method to a larger number of pseudo-competitors in Appendix~\ref{appendix:rq_4_synthetic}.} 
We are aware of the cases that collecting corpora in target languages for competitions might be significantly complex, which means only a monolingual corpus is available for evaluation.  
Thus, we train predictors $f'$ using single \selfscores in Equation~\ref{eq:single_direct}.
Note that this experiment covers several challenging settings, such as transferred MT systems, language transferability, single source features, and transferred application domains. Another set of results on \wmtbio can be found in Appendix~\ref{appendix:rq_4_bio}.

\paragraph{Discussion.} In Table~\ref{tab:wmt20news_results}, we display the results on \wmtnews.
Although MAE and RMSE vary among experiments for different language pairs, the overall correlation scores are favorable. When using $B \circlearrowright A$ features or both $A \circlearrowright B$ and $B \circlearrowright A$ features, Pearson's $r$ values on all language pairs are no lower than 0.5, showing strong ranking correlations. While prediction performances on $A \circlearrowright B$ have some variances among different language pairs, the results of the experiments using $B \circlearrowright A$ are competitive to those using both $A \circlearrowright B$ and $ B \circlearrowright A$ features, showing the feasibility of predicting \transscore using monolingual data. We conclude that our regression-based predictors can be practical in ranking MT systems in \wmt-style shared tasks.


%% file: tables/exp_system_transfer.tex
\begin{table}[h]
\centering
\resizebox{1.0\linewidth}{!}{
\begin{tabular}{lc|ccc}
\hline
\multirow{2}{*}{\textbf{MT System}} 
& \multirow{2}{*}{\textbf{Trans. Metric}} 
& \multicolumn{3}{c}{\textbf{Type I}} 

\\ \cline{3-5} 
& & \textbf{MAE} $\downarrow$ & \textbf{RMSE} $\downarrow$ & \textbf{P.} 
$\boldsymbol r$ $\uparrow$

\\ \hline
\cline{1-5}
\multirow{4}{*}{\mbart}
&\sacrebleu & 1.80 & 2.70 & 0.94 \\
&\spbleu & 2.13 & 2.99 & 0.94\\
&\chrf & 3.51 & 4.53 & 0.96\\
&\bertscore & 4.98 & 7.07 & 0.88\\

\cline{1-5}
\multirow{4}{*}{\mmbase}
&\sacrebleu &3.86 & 5.82 & 0.95 \\
&\spbleu &3.97 & 5.72 & 0.96\\
&\chrf & 6.06 & 7.53 & 0.96\\
&\bertscore & 4.35 & 6.32 & 0.91\\
\cline{1-5}
\multirow{4}{*}{\googletrans}
&\sacrebleu &4.09 & 5.60&0.93\\
&\spbleu &4.22 & 5.62&0.87\\
&\chrf & 5.70& 6.90&0.93\\
&\bertscore & 2.87& 3.66&0.80\\
\hline
\end{tabular}
}
\caption{The results of predicted \transscores of \mbart, \mmbase and \googletrans on \typeI test set based on different translation evaluation metrics (Trans. Metric). *MAE: Mean Absolute Error, RMSE: Root Mean Square Error, P. $\boldsymbol r$: Pearson's $r$.}
\label{tab:system_transfer}
\vspace{-6mm}
\end{table}

%% file: tables/exp_language_transfer.tex
\begin{table}[h]
\small
\centering
\resizebox{1.0\linewidth}{!}{
\begin{tabular}{lc|ccc|ccc}
\hline
\multirow{2}{*}{\textbf{MT System}} 
& \multirow{2}{*}{\textbf{Trans. Metric}} 
& \multicolumn{3}{c|}{\textbf{Type II}} 
& \multicolumn{3}{c}{\textbf{Type III}} 

\\ \cline{3-8} 
& & \textbf{MAE} $\downarrow$ & \textbf{RMSE} $\downarrow$ & \textbf{P.} 
$\boldsymbol r$ $\uparrow$
& \textbf{MAE} $\downarrow$ & \textbf{RMSE} $\downarrow$ & \textbf{P.} 
$\boldsymbol r$ $\uparrow$

\\ \hline
\multirow{4}{*}{\mbart}
&\sacrebleu & 1.36& 1.97& 0.93& 0.81& 0.95&0.96\\
&\spbleu & 1.61 & 2.19& 0.93& 1.20&1.38 &0.94\\
&\chrf & 3.80& 4.89& 0.95& 3.04& 3.89&0.95\\
&\bertscore & 4.67& 6.38& 0.88& 5.08& 6.88&0.87\\
\cline{1-8}
\multirow{4}{*}{\mmbase}
&\sacrebleu & 3.10& 4.16& 0.95& 2.99& 3.76&0.94\\
&\spbleu & 3.24 & 4.18& 0.96& 3.18&3.88 &0.95\\
&\chrf & 5.53& 6.70& 0.95& 5.42& 6.54&0.93\\
&\bertscore & 4.38& 6.51& 0.83& 4.29& 6.65&0.80\\
\hline
\end{tabular}
}
\caption{The results of predicted \transscores of \mbart (a \textit{seen} MT system) and \mmbase (an \textit{unseen} MT system) on \typeII and \typeIII (with \textit{unseen} languages) test sets based on different translation evaluation metrics (Trans. Metric).}
\label{tab:language_transfer}
\vspace{-3mm}
\end{table}

%% file: tables/wmt20_news_results.tex
\begin{table*}[h]
\centering
\small
\resizebox{0.9\linewidth}{!}{
\begin{tabular}{c|cccc|cccc|cccc}
\hline
\multirow{2}{*}{\textbf{Lang. Pair}} 
& \multicolumn{4}{c|}{$A \circlearrowright B$} & \multicolumn{4}{c|}{$B \circlearrowright A$} &\multicolumn{4}{c}{$A \circlearrowright B$ \& $ B \circlearrowright A$}\\
\cline{2-13}
& \textbf{MAE} $\downarrow$ & \textbf{RMSE} $\downarrow$ &\textbf{K.} $\boldsymbol \tau$ $\uparrow$& \textbf{P.} $\boldsymbol r$ $\uparrow$ 
& \textbf{MAE} $\downarrow$ & \textbf{RMSE} $\downarrow$ &\textbf{K.} $\boldsymbol \tau$ $\uparrow$& \textbf{P.} $\boldsymbol r$ $\uparrow$
& \textbf{MAE} $\downarrow$ & \textbf{RMSE} $\downarrow$ &\textbf{K.} $\boldsymbol \tau$ $\uparrow$&\textbf{P.} $\boldsymbol r$ $\uparrow$\\
\hline
cs-en & 4.01 & 4.34 & 0.20 & 0.45&8.92 & 9.08&0.60 & 0.91 &8.53 &8.71 &0.60 &0.88\\
de-en & 13.23 & 13.26 & 0.80 & 0.95&1.69 & 1.77 &0.80 & 0.95 &1.26&1.38&0.80&0.96\\
de-fr & 10.45 & 10.53 & 1.00 & 0.99& 1.72& 2.05&0.80 & 0.97 &1.59&1.93 & 1.00&0.97\\
en-cs & 6.96 & 7.49 & 0.20 & 0.25& 1.39& 1.79&0.60 & 0.94 &1.25 &1.80&0.60&0.95\\
en-de & 2.96 & 4.00 & 0.40 & 0.59& 2.29& 2.70&1.00 & 0.92 &2.75 &3.12&1.00&0.93\\\
en-ru & 1.98 & 2.40 & 0.20 & 0.40&7.41 & 7.53&0.40 & 0.85 &7.48 &7.60&0.60 &0.86\\
en-zh & 2.96 & 3.93 & 0.20 & 0.19& 1.36& 1.60& 0.80 & 0.80 &1.23 &1.50&0.80&0.82\\
fr-de & 2.89 & 3.70 & 0.80 & 0.90& 2.99& 3.56& 1.00 & 0.94 &2.59 &3.17&1.00&0.93\\
ru-en & 9.83 & 9.97 & 1.00 & 0.78&1.16 & 1.72&0.80 & 0.85 &1.44&1.78&0.80&0.88\\
zh-en & 12.44 & 12.77 & 0.00 & 0.26& 3.04& 3.55& 0.20 & 0.50 &2.62 &3.56 &0.20 &0.50\\
\hline
Average & 6.77 & 7.24 & 0.48 & 0.58 & 3.20&3.54 & 0.70 & 0.86 &3.07&3.46&0.74&0.87\\
\hline
\end{tabular}
}
\caption{The results of our predictors on ranking the selected MT systems on \wmtnews shared tasks.}
\label{tab:wmt20news_results}
\centering
\vspace{-6mm}
\end{table*}

%% file: 07_3_exp_improve_QE.tex
\input{tables/da20_qe}
\input{tables/wmt20_cross_check}

\subsubsection{\selfscores for Quality Estimation}
In this section, we demonstrate that the features acquired by round-trip translation benefit quality estimation (QE) models.


\paragraph{Dataset.}
QE was firstly introduced in WMT11 \cite{callison2011findings}, focusing on automatic methods for estimating the quality of neural machine translation output at run-time. The estimated quality should align with the human judgment on the word and sentence level, without accessing to the reference in the target language. In this experiment, we perform sentence-level QE, which aims to predict human direction assessment (DA) scores. We use DA dataset collected from 2015 to 2021 by MT News Translation shared task coordinators. More details are provided in Appendix~\ref{appendix:qe_dataset}.

\paragraph{Settings.}

Firstly, we extract \rtt features \rtt-\sacrebleu, \rtt-\spbleu, \rtt-\chrf and \rtt-\bertscore. Then, we examine whether QE scores could be predicted by these \rtt features using linear regression models. We train the regressors using Equation~\ref{eq:single_direct} with only $A\circlearrowright B$ features. Finally, a combination of \cometda scores and \selfscores are investigated to acquire a more competitive QE scorer.


\paragraph{Discussion.} Both Kendall's $\tau$ and Pearson's $r$ provide consistent results in Table~\ref{tab:da_qe}. The models merely using \selfscores could be used to predict DA scores. We also observe that \selfscores can further boost the performance of \cometda. 
We believe \selfscores advances QE research and urge more investigation in this direction.



%% file: tables/da20_qe.tex
\begin{table}[!ht]
    \centering
    \small
    \centering
    \begin{tabular}{l|cc|cc}
    \hline
    \multirow{2}{*}{QE model} &   \multicolumn{2}{c|}{zh-en} & \multicolumn{2}{c}{en-de} \\\cline{2-5}
    & \textbf{K.}  $\boldsymbol \tau$ $\uparrow$ & \textbf{P.}  $\boldsymbol r$ $\uparrow$ & \textbf{K.}  $\boldsymbol \tau$ $\uparrow$ & \textbf{P.}  $\boldsymbol r$ $\uparrow$ \\\hline
    \rtt-\sacrebleu  & 15.17& 21.76 & 11.83& 19.71\\
    \rtt-\spbleu & 13.55 & 18.30 & 11.49 & 19.00\\
    \rtt-\chrf & 15.52 & 21.74 & 13.57& 22.93\\
    \rtt-\bertscore & 15.70 & 21.96 & 25.89 & 44.10\\
    \rtt-ALL & 15.90 & 22.36 & 26.02 & 44.33\\ 
    \hline
    \cometda  & 32.83 & 46.91 & 42.71 & 64.36 \\
     \ \ \ \ + \rtt-ALL &\textbf{33.52} & \textbf{47.88} & \textbf{44.23} & \textbf{66.74}\\\hline
    \end{tabular}
    \caption{Comparisons of \selfscore for QE. \rtt-ALL refers to the combination of \rtt-\sacrebleu, \rtt-\spbleu and \rtt-\chrf. \cometda + \rtt-ALL incorporates both \cometda and all \selfscores.}
    \label{tab:da_qe}
\vspace{-5mm}
\end{table}

%% file: tables/wmt20_cross_check.tex
\begin{table*}[!ht]
\small
    \centering
    \resizebox{1.00\linewidth}{!}{
    \begin{tabular}{cc|cc|cccc|cccc}
    \hline
        \multirow{2}{*}{\textbf{\# Sys.}} & \multirow{2}{*}{Method} & \multicolumn{2}{c|}{No Adversary} & \multicolumn{4}{c|}{One adversarial SMT} &\multicolumn{4}{c}{Two adversarial SMTs}\\
        \cline{3-12}
        & & \textbf{K.}  $\boldsymbol \tau$ $\uparrow$& \textbf{P.}  $\boldsymbol r$ $\uparrow$ 
& \textbf{Hit@1} $\uparrow$ & \textbf{Avg. Rank} $\downarrow$ &\textbf{K.}  $\boldsymbol \tau$ $\uparrow$& \textbf{P.}  $\boldsymbol r$ $\uparrow$
& \textbf{Hit@2} $\uparrow$ & \textbf{Avg. Rank} $\downarrow$ &\textbf{K.}  $\boldsymbol \tau$ $\uparrow$& \textbf{P.}  $\boldsymbol r$ $\uparrow$\\\hline
         \multirow{2}{*}{3}
         & \xmark & 0.07 & 0.17& 0.50 & 2.00 & 0.33 & 0.51& 0.00 & 4.75 & -0.15 & -0.30\\
         &\cmark & 0.47 & 0.43 & 1.00 & 1.00 & 0.33 & 0.98& 1.00 & 1.50 & 0.55 & 0.98\\\hdashline
         \multirow{2}{*}{4}
         &\xmark& 0.33 & 0.37& 0.25 & 2.75 & 0.40 & 0.39& 0.00 & 5.75 & -0.03 & -0.33\\
         &\cmark& 0.57 & 0.81& 1.00 & 1.00 & 0.60 & 0.97& 1.00 & 1.50 & 0.70 & 0.98\\\hdashline
         \multirow{2}{*}{5}
         &\xmark& 0.48 & 0.58& 0.25 & 3.25 & 0.30 & 0.25& 0.00 & 6.75 & -0.05 & -0.40\\
         &\cmark& 0.42 & 0.52& 1.00 & 1.00 & 0.50 & 0.93& 1.00 & 1.50 & 0.62 & 0.92\\\hline
    \end{tabular}}
    \caption{Results of the competition between 3 to 5 honest competitors, with a combination of additional adversarial competing systems (No Adversary, One adversarial SMT ($X=0.1$) w/ copy, Two adversarial SMTs ($X=0.1$ and $X=0.5$) w/ copy). We measure the identifiability of the adversarial MT systems by Hit@$K$, where $K$ is decided by the number of adversarial systems. We also report the average ranking (Avg. Rank.) of the adversarial systems, and correlation scores, Kendall's $\tau$ and Pearson's $r$.}
    \label{tab:xcheck}
    \vspace{-3mm}
\end{table*}

%% file: 07_2_exp_cross_system.tex
\section{Towards Robust Evaluation}
On the basis of our findings in Section~\ref{sec:smt_exp}, \rtt evaluation could become potentially vulnerable when MT systems with word copy are involved in. Specifically, the adversarial system may achieve unexpectedly high \selfscores due to the large portion of preserved words inside original context via \rtt, while its \transscores remain low.

In order to mitigate the vulnerability, we first validate \rtt evaluation on \wmtnews with $A \circlearrowright B$ direction. One of the advantages of \rtt is that multiple MT systems could be used to verify the performance of other systems via checking the $N\times N$ combinational \rtt results from these $N$ systems, coined \xcheck. 
Finally, we demonstrate that the predicted automatic evaluation scores could be further improved via \xcheck when adversaries are included. 

\subsection{Cross-system Validation for Competitions}
Given \ft MT systems $\{\mathcal F_i\}_{i=1}^{N}$, \bt MT systems $\{\mathcal B_i\}_{i=1}^{M}$, and a regression model $\mathcal{M}$ on predicting the target metric, we can estimate the translation quality of $i$-th \ft system on $j$-th \bt system:

\begin{align} \nonumber
    \mathbb{S}_{i,j} & =f_\mathcal{M} (\mathcal{B}_j(\mathcal{F}_i(x)), x),
\end{align}
where $\mathbb{S} = \{\mathbb{S}_{i,j}\}_{N\times M} $.
The estimated translation quality of $\mathcal{F}_i$ is the average score of the $i$-th column,
$$
    \overline{\mathbb{S}}_{i, :} = \frac{1}{M} \sum_{j=1}^{M} \mathbb{S}_{i,j}.
$$

Note that the same number of \ft and \bt systems are considered for simplicity, i.e. $N=M$.  

\subsection{Experiments and Analysis}

\paragraph{Settings.} We conduct experiments on \wmtnews similar to Section~\ref{rq3}. 
We rank the system-level translation quality via the regressor trained on $\text{\selfscore}^\spbleu$. We challenge the evaluation paradigm by introducing some adversarial MT systems, e.g., SMT with copying mechanism. 
Specifically, we introduce basic competition scenarios with 3-5 competitors 
to the shared task, and we consider different numbers of adversarial systems, namely \textit{i)} no adversary; \textit{ii)} one adversarial SMT with word copy; \textit{iii)} two adversarial SMT systems with word copy. We provide details of two SMT systems in Appendix~\ref{appendix:xcheck}. 
The experiments with adversarial systems are conducted on four language pairs, $\texttt{cs-en}$, $\texttt{de-en}$, $\texttt{en-cs}$ and $\texttt{en-de}$, as the corresponding adversarial systems were trained in Section~\ref{sec:smt_vs_nmt_exp}. 

\paragraph{Discussion.} From Table~\ref{tab:xcheck}, we observe that the overall system ranking could be severely affected by the adversarial systems, according to Pearson's $r$ and Kendall's $\tau$. The adversarial systems are stealthy among normal competitors, according to Hit@K and Avg. Rank.
\xcheck evidently successfully identifies these adversarial systems in all our experiments and manages to improve the correlation scores significantly. With the empirical study, we find that \xcheck is able to make \rtt evaluation more robust.


%% file: 10_conclusion.tex
\section{Conclusion} 
This paper revisits the problem of estimating \ft quality using \rtt scores. The negative results from previous literature are basically caused by the heavy reliance of copy mechanism in traditional statistical machine translation systems. 
Then, we conduct comprehensive experiments to show the corrected understanding on \rtt benefits several relevant MT evaluation tasks, such as predicting \ft metrics using \rtt scores, enhancing state-of-the-art QE systems, and filtering out unreliable MT competitors for \wmt shared tasks. We believe our work will inspire research on reference-free evaluation on low-resource machine translation.


 
\section*{Limitations} 

Although we have observed positive correlation between \transscores and \selfscores and conducted experiments to predict \transscores using \selfscores, their relations could be complicated and non-linear.
We encourage future research to investigate various \selfscore features and more complex machine learning models for better prediction models. 
Although we have examined the prediction models on low-resource languages in \flores, we have not tested those very low-resource languages out of these 101 languages. We suggest auditing \transscore prediction models on a small validation dataset for any new low-resource languages in future applications. 

%% file: 08_appendix.tex

\section{Dataset Construction}
\label{appendix:dataset_detail}
\input{tables/selected_flores_language}

We provide the statistics of all languages covered by \dataset, categorised by different scale of the resource (high, medium and low) and usage purpose (\seen and \unseen) in Table~\ref{tab:flores_ae33}.  Scale is counted by the amount of bi-text data to English in \flores~\citep{flores101}.

To construct \dataset, we partition these 33 languages into two sets, \textit{i)} the languages that are utilized in training our models (\seen\footnote{Both train and test sets of our corpus will have these languages.}) and \textit{ii)} the others are employed used for training the predictors but considered for test purpose only (\unseen).
We include 20 languages to \seen, with 7 high-resource, 7 medium-resource and 6 low-resource. The rest 13 languages fall into \unseen, with 9 medium-resource and 4 low-resource. Combining these two categories of languages, we obtain three types of \textit{language pairs} in \dataset.

\typeI contains pairs of languages in \seen, where a train set and a test set are collected and utilized independently. For each language pairs, we collect 997 training samples and 1,012 test samples.
The test set of \typeII is more challenging than that of \typeI set, where the language pairs in this set are composed of one language from \seen set and the other language from \unseen set. 
\typeIII's test set is the most challenging one, as all its language pairs are derived from \unseen languages. \typeII and \typeIII sets are designed for test purpose, and they will not be used for training predictors. Overall, \typeI, \typeII and \typeIII sets contain 380, 520, and 156 language pairs, respectively.

\section{Automatic Evaluation Metrics for Translation}
\label{appendix:evaluation_metrics}
For \bertscore, Deberta-xlarge-mnli~\cite{he2021deberta} is used as the backbone pre-trained language model, as it is reported to have a satisfactory correlation with human evaluation in WMT16.
While \sacrebleu, \spbleu and \chrf are string-based metrics, \bertscore is model-based. The selection of these metrics is on the basis that they should directly reflect the translation quality. We calculate those scores via open-source toolboxes, \easynmt\footnote{\url{https://github.com/UKPLab/EasyNMT}.}, \bleutool\footnote{\url{https://github.com/mjpost/sacrebleu}.} and \berttool\footnote{\url{https://github.com/Tiiiger/bert_score}.}. We use word-level 4-gram for \sacrebleu and \spbleu, character-level 6-gram for \chrf, and $F_1$ score for \bertscore by default. 
\section{Machine Translation Systems}
\label{appendix:mts_detail}
\paragraph{\moses SMT.} We train five Moses'~\cite{koehn2009moses} statistical machine translation systems using different phrase dictionary by varying phrase probability threshold from 0.00005, to 0.5. 
The higher threshold indicates the smaller phrase table and hence a better chance of processing unknown words by the corresponding MT systems. 
In Table~\ref{tab:word_copy}, we use \moses with the phrase probability threshold of 0.4 for SMT.
\paragraph{\mbart.} \mbart~\cite{tang2020multilingual} is a multilingual translation model with many-to-many encoders and decoders. The model is trained on 50 publicly available language corpora with English as a pivot language.
\paragraph{\mmbase \& \mmlarge.} These two models are one of the first non-English-centric multilingual machine translation systems, which are trained on 100 languages covering high-resource to low-resource languages. Different from \mbart, \mmbase and \mmlarge~\cite{fan2021beyond} are trained on parallel multilingual corpora without an explicit centering language.

\paragraph{\opus.} \opus~\cite{tiedemann2020opus} is a collection of one-to-one machine translation models which are trained on corresponding parallel data from OPUS using \marian as backbone~\cite{junczys2018marian}. The collection of MT models support 186 languages.

\paragraph{\googletrans.} \googletrans~\cite{wu2016google, bapna2022building} is an online Translation service provided by Google Translation API, which supports 133 languages. The system is frequently involved as a baseline system by \wmt shared tasks~\cite{barrault-etal-2020-findings}.

\section{Quality Estimation Dataset}
\label{appendix:qe_dataset}
The direct-assessment (DA) \textit{train set} contains 33 diverse language pairs and a total of 574,186 tuples with source, hypothesis, reference and direct
assessment z-score. We construct the \textit{test} set by collecting DA scores on \textit{zh-en} (82,692 segments) and \textit{en-de} (65,045 segments), as two \textit{unseen} language pairs.

\section{Implementation Details}
\label{appendix:implement}
\paragraph{Regressor.} We use the linear regression model tool by \texttt{Scikit-Learn}\xspace\footnote{\url{https://scikit-learn.org/stable/modules/generated/sklearn.linear_model.LinearRegression.html}} with the default setting for the API.
\paragraph{MT Systems.} We adopt \texttt{EasyNMT}\xspace\footnote{\url{https://github.com/UKPLab/EasyNMT}} for loading \mbart, \mmbase, \mmlarge and \opus for translation.

\paragraph{Computational Resource and Time.} In our experiment, we collect the translation results and compute their \transscore and \selfscore on multiple single-GPU servers with Nvidia A40. Overall, it cost us about three  GPU months for collecting translation results by all the aforementioned MT systems.

\input{figures/all_correlation_vis}

\section{Measurement}
\label{appendix:measurement_detail}
We evaluate the performance of our predictive model via the following measurements:
\paragraph{Mean Absolute Error (MAE)} is used for measuring the average magnitude of the errors in a set of predictions, indicating the accuracy for continuous variables.

\paragraph{Root Mean Square Error (RMSE)} measures the average magnitude of the error. Compared to MAE, RMSE gives relatively higher weights to larger error.

\paragraph{Pearson's $\boldsymbol r$ correlation}~\cite{benesty2009pearson} is officially used in \wmt to evaluate the agreement between the automatic evaluation metrics and human judgement, emphasizing on the translation consistency. In our paper, the metric evaluates the agreement between the predicted automatic evaluation scores and the ground truth.

\paragraph{Kendall's $\boldsymbol \tau$ correlation}~\cite{kendall1938new} is another metric to evaluate the ordinal association between two measured quantities. 

\section{Supplementary Experiments}

\subsection{Correlation between \transscores and \selfscores on \dataset}
\label{appendix:rq_1}
\input{tables/pearson_r_correlations}
\paragraph{Settings.} We experiment with \mbart and  \mmbase on \textbf{Type I} test set of \dataset by comparing their $\selfscore_{A \circlearrowright B}^{\mathcal{M}}$, $\selfscore_{B \circlearrowright A}^{\mathcal{M}}$ and $\transscore_{A \rightarrow B}^{\mathcal{M}}$ using multiple translation metrics $\mathcal M$, \sacrebleu, \spbleu, \chrf and \bertscore.
We measure their correlations by computing Pearson's $r$~\cite{benesty2009pearson} of $(\selfscore_{A \circlearrowright B}^{\mathcal M}, \transscore_{A \rightarrow B}^{\mathcal M})$ and $(\selfscore_{B \circlearrowright A}^{\mathcal M}, \transscore_{A \rightarrow B}^{\mathcal M})$. Note that our experiment is beyond English-centric, as all languages are permuted and equally considered.

\input{tables/ablation}
\paragraph{Discussion.} The overall correlation scores are reported in Table~\ref{tab:correlation_all}. Our results indicate at least moderately positive correlations between all pairs of \selfscores and \transscores. Moreover, we observe that $\selfscore_{B \circlearrowright A}$ is generally more correlated to \transscore than $\selfscore_{A \circlearrowright B}$, leading to strongly positive correlation scores. We attribute the advantage to the fact that $\mathcal{T}_{A\rightarrow B}$ serves as the last translation step in $\selfscore_{B \circlearrowright A}$. 
We visualize more detailed results of correlation between \transscores and \selfscores on \typeI language pairs in \flores, in Figure~\ref{fig:mbart} (\mbart) and Figure~\ref{fig:mmbase} (\mmbase).

\subsection{Improve Prediction Performance Using More Features}
\label{appendix:more_features}
\paragraph{Settings.} We introduce two extra features, \MaxCount and \RefLen,\footnote{\MaxCount and \RefLen are ``counts'' and ``ref\_len'' in \url{https://github.com/mjpost/sacrebleu/blob/master/sacrebleu/metrics/bleu.py}.} to enhance the prediction of \spbleu.  \MaxCount is the counts of correct 4 grams, and \RefLen is the cumulative reference length. We follow the similar procedure in \textbf{RQ2}, using the same measurements to evaluate the predictor performance on \mbart and \mmbase across three types of test sets in \dataset.

\paragraph{Results.} Table~\ref{tab:rq_5} shows the results of those models with additional features. 
Both features consistently improve our basic models, and 
the performance can be further boosted by incorporating both features. We believe that more carefully designed features and regression models could potentially boost the performance of our predictors. 

\subsection{\wmtnews with Synthetic Competitors}
\label{appendix:rq_4_synthetic}

We increase the scale of competitors to \wmtnews by introducing pseudo competitors. To mimic the number of a conventional \wmt task, we vary 17 forward translation systems by randomly dropping $0\%$ to $80\%$ (with a step of $5\%$) tokens from the outputs of \googletrans. Then, we utilize the vanilla \googletrans to translate these synthetic forward translation results back to the source language.
We conduct experiments on \textit{de-fr}, \textit{en-ta} and \textit{zh-en}, representing those \textit{non-En to non-En}, \textit{En to non-En} and \textit{non-En to En}  language pairs.  

The results in Table~\ref{tab:rq4_synthetic_data} demonstrate the predictors' performances on ranking the pseudo competitors on \wmtnews based on \spbleu features. The overall ranking errors on 17 MT systems are small on all three selected language pairs. 

\input{tables/synthetic_results}

\subsection{Ranking Experiments on \wmtbio}
\label{appendix:rq_4_bio}

We display the experimental results on \wmtbio in the Table \ref{tab:wmt20bio_results}. The overall performance is positive, while it is relatively worse than the results of \wmtnews reported in Table \ref{tab:wmt20news_results}. We attribute this to the fact that the $\mathcal M$ used on \wmtbio are calculated on document, while our regression models rely on sentence-level translation metrics in training. The large granularity difference of text may result in a distribution shift.  
\input{tables/wmt20_bio}



\subsection{Benign MT systems and Adversarial MT Systems for \xcheck}
\label{appendix:xcheck}

The selection of the benign systems is: 
\begin{itemize}
    \item \textbf{3 Systems:} \opus, \mmlarge and \mbart; 
    \item \textbf{4 Systems:} \opus, \mmlarge,\mmbase and \mbart; 
    \item \textbf{5 Systems:} \googletrans, \opus, \mmlarge,\mmbase and \mbart.
\end{itemize}

\paragraph{SMT ($X=0.1$).} We train the SMT system on \newscommentary with the max phrase length 4 and the phrase table probability threshold of 0.1.
\paragraph{SMT ($X=0.5$).} We train the SMT system on \newscommentary with the max phrase length 4 and the phrase table probability threshold of 0.5.

SMT($X=0.1$) tends to copy fewer words than SMT($X=0.5$), due to the larger phrase table size filtered by lower prebability threshold.

%% file: tables/selected_flores_language.tex
\begin{table}[!ht]
    \centering
\resizebox{1.0\linewidth}{!}{
    \begin{tabular}{c|cccc}
        \hline
      \textbf{Resource} &  \textbf{Language} &  \textbf{Scale} & \textbf{Usage}\\ \hline
         \multirow{7}{*}{High} 
         & English & - & \seen\\
         & Spanish& 315M & \seen\\
         & French & 289M & \seen\\
         & German& 216M & \seen\\
         & Portuguese& 137M & \seen\\
         & Russian& 127M & \seen\\
         & Italian& 116M & \seen\\
         
         \hline
         \multirow{16}{*}{Medium} 
         & Dutch& 82.4M & \seen\\
         & Turkish& 41.2M & \seen\\
         & Polish& 40.9M & \seen\\
         & Chinese& 37.9M & \seen\\
         & Romanian& 31.9M & \seen\\
         & Greek& 23.7M & \seen\\
         & Japanese & 23.2M & \seen\\
         & Czech & 23.2M & \unseen\\
         & Finnish & 15.2M & \unseen\\
         & Bulgarian&10.3M & \unseen\\
         & Lithuanian&6.69M & \unseen\\
         & Estonian&4.82M & \unseen\\
         & Latvian&4.8M & \unseen\\
         &Hindi&3.3M & \unseen\\
         &Javanese&1.49M & \unseen\\
         &Icelandic&1.17M & \unseen\\
         \hline
         \multirow{10}{*}{Low} 
         & Tamil & 992K & \seen\\
         & Armenian & 977K & \unseen\\
         & Azerbaijani & 867K & \unseen\\
         & Kazakh & 701K & \seen\\
         & Urdu&630K & \unseen\\
         & Khmer & 398K & \seen\\
         & Hausa & 335K & \seen\\
         & Pashto & 293K & \seen\\
         & Burmese & 283K & \unseen\\
         & Gujarati & 160K & \seen\\
         \hline
    \end{tabular}
}
    \caption{The statistics of \dataset. 
    20 languages are used in both training and test (\seen), the other 13 languages are used in test only (\unseen).}
    \label{tab:flores_ae33}
\end{table}

%% file: figures/all_correlation_vis.tex
\begin{figure*}[!htb]
\centering
\begin{tabular}{cccc}
\subcaptionbox{\label{a1}}{\includegraphics[width=0.2\linewidth]{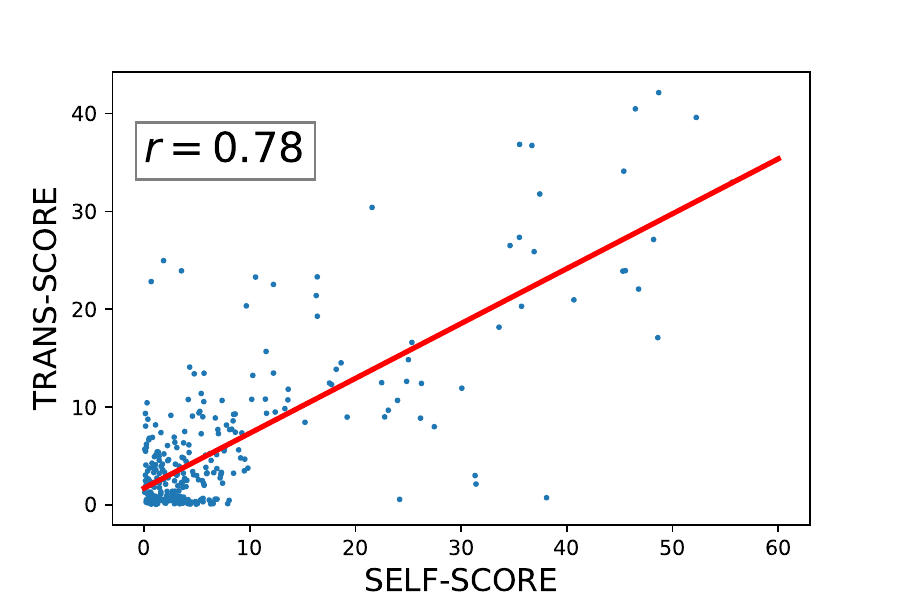}} &
\subcaptionbox{\label{b1}}{\includegraphics[width=0.2\linewidth]{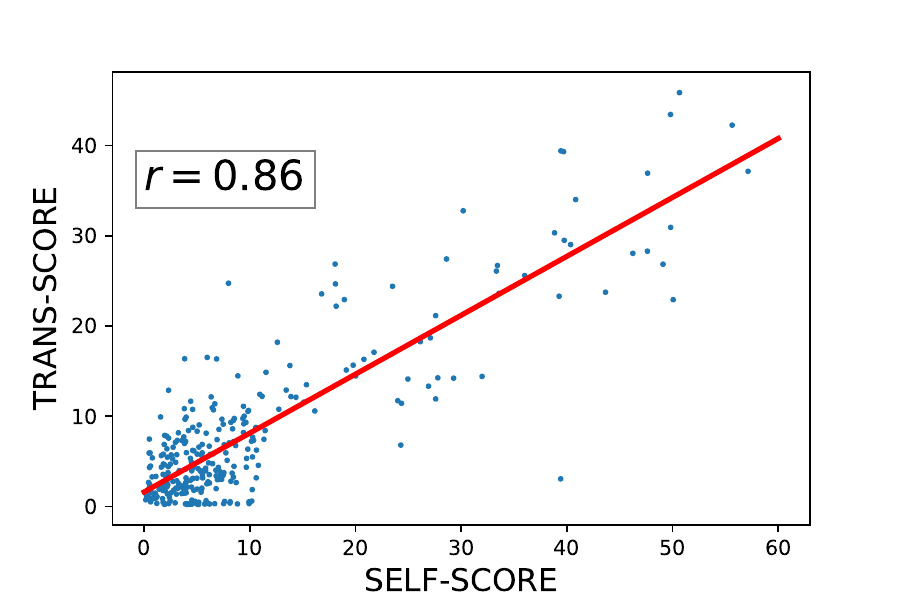}} 
&\subcaptionbox{\label{c1}}{\includegraphics[width=0.2\linewidth]{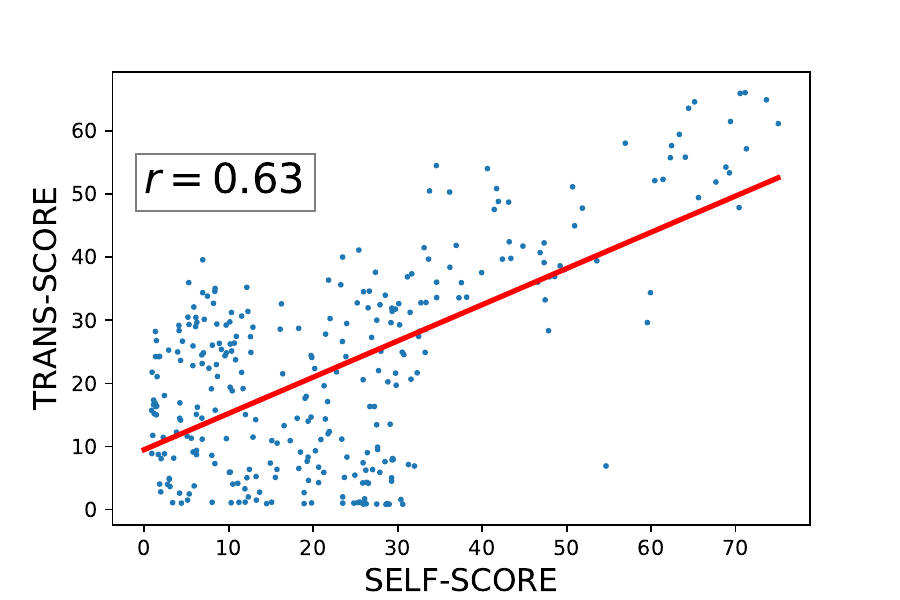}} &
\subcaptionbox{\label{d1}}{\includegraphics[width=0.2\linewidth]{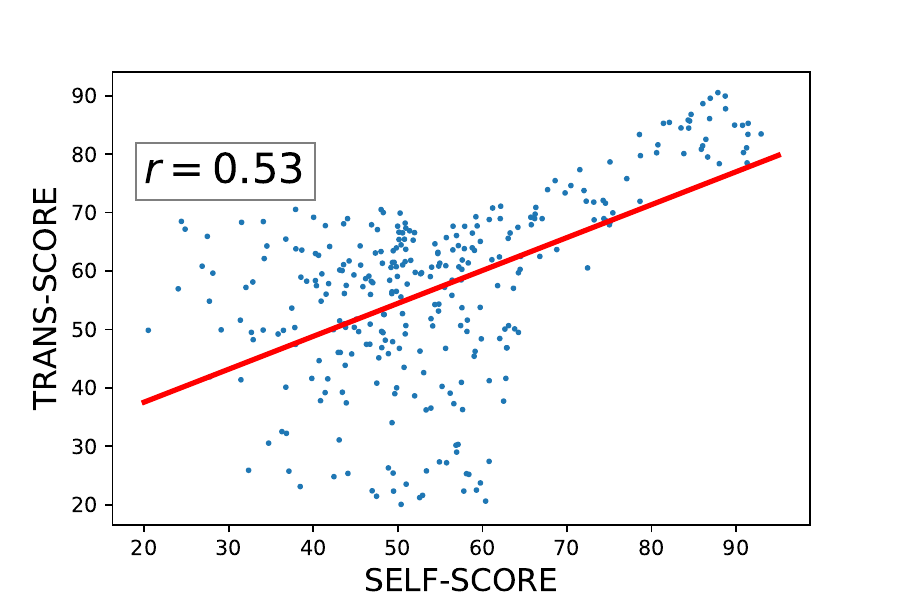}}
\\
\subcaptionbox{\label{a2}}{\includegraphics[width=0.2\linewidth]{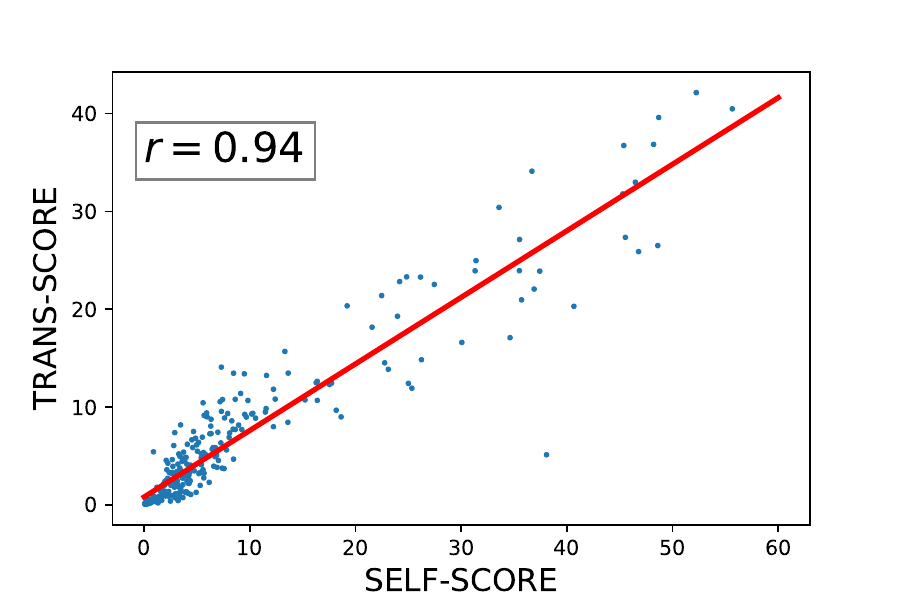}} &
\subcaptionbox{\label{b2}}{\includegraphics[width=0.2\linewidth]{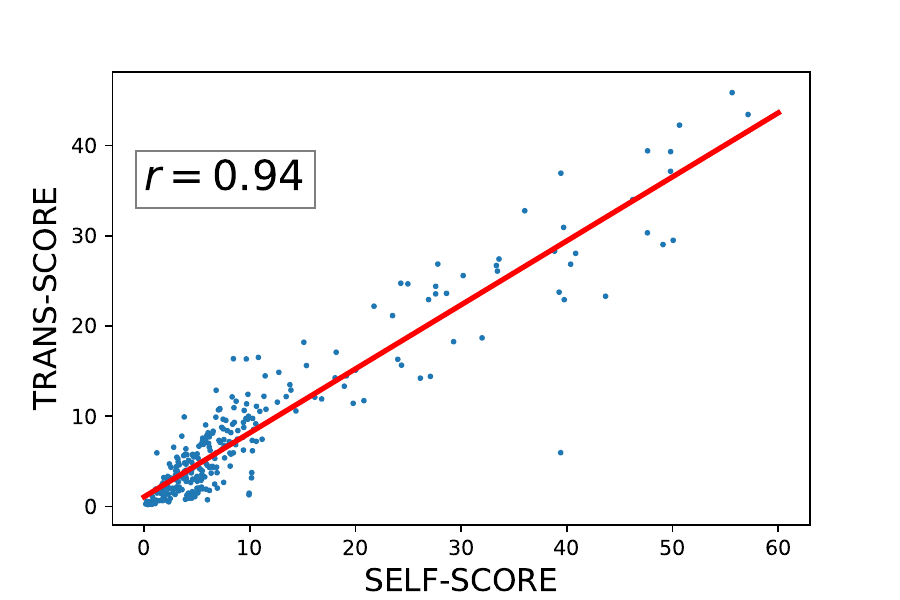}} 
&\subcaptionbox{\label{c2}}{\includegraphics[width=0.2\linewidth]{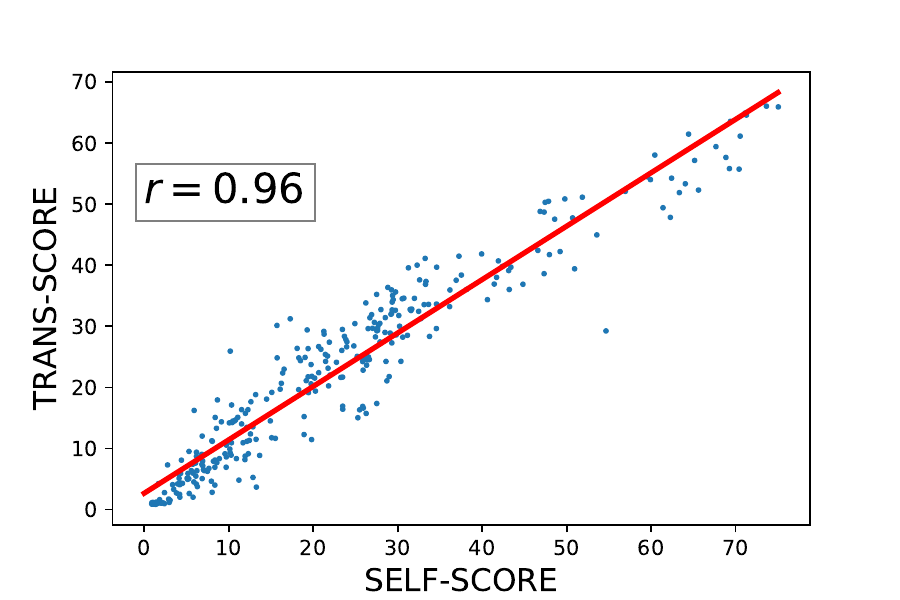}} &
\subcaptionbox{\label{d2}}{\includegraphics[width=0.2\linewidth]{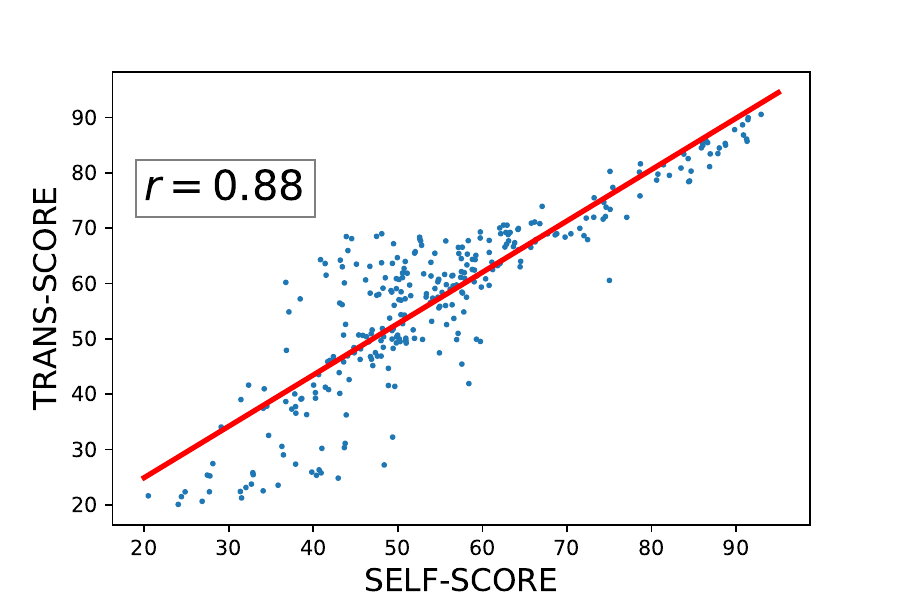}}\\
\end{tabular}
\caption{\protect\raggedright The first row is the correlations between $\text{\selfscore}_{A\circlearrowright B}^{\mathcal{M}}$ and $\text{\transscore}_{A\rightarrow B}^{\mathcal{M}}$ on \mbart using (a) \sacrebleu, (b) \spbleu, (c) \chrf and (d) \bertscore. The second row is the correlations between $\text{\selfscore}_{B\circlearrowright A}^{\mathcal{M}}$ and $\text{\transscore}_{A\rightarrow B}^{\mathcal{M}}$ on \mbart using (e) \sacrebleu, (f) \spbleu, (g) \chrf and (h) \bertscore.  All experiments with overall Pearson's $r$.}
\label{fig:mbart}
\end{figure*}

\begin{figure*}[!htb]
\centering
\begin{tabular}{cccc}
\subcaptionbox{\label{a}}{\includegraphics[width=0.2\linewidth]{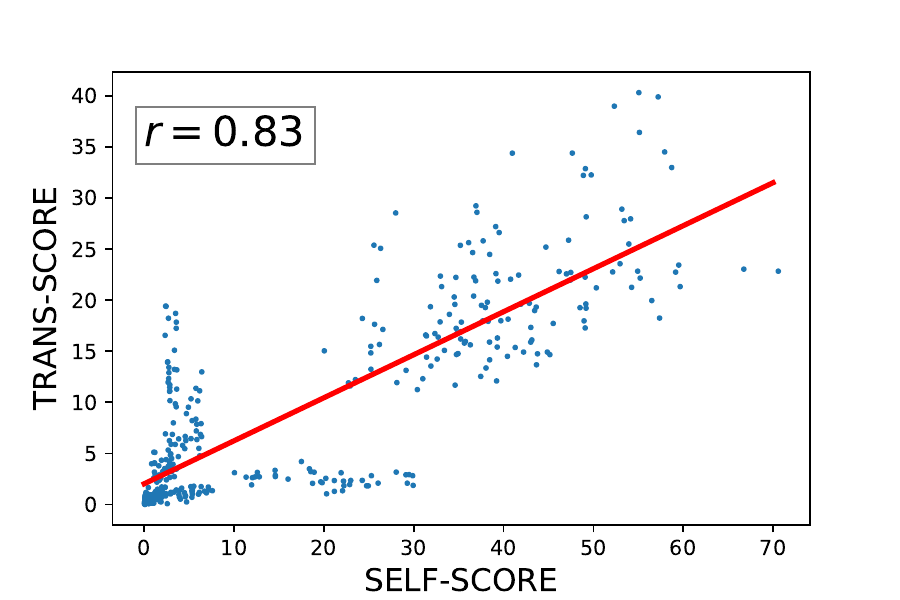}} &
\subcaptionbox{\label{b}}{\includegraphics[width=0.2\linewidth]{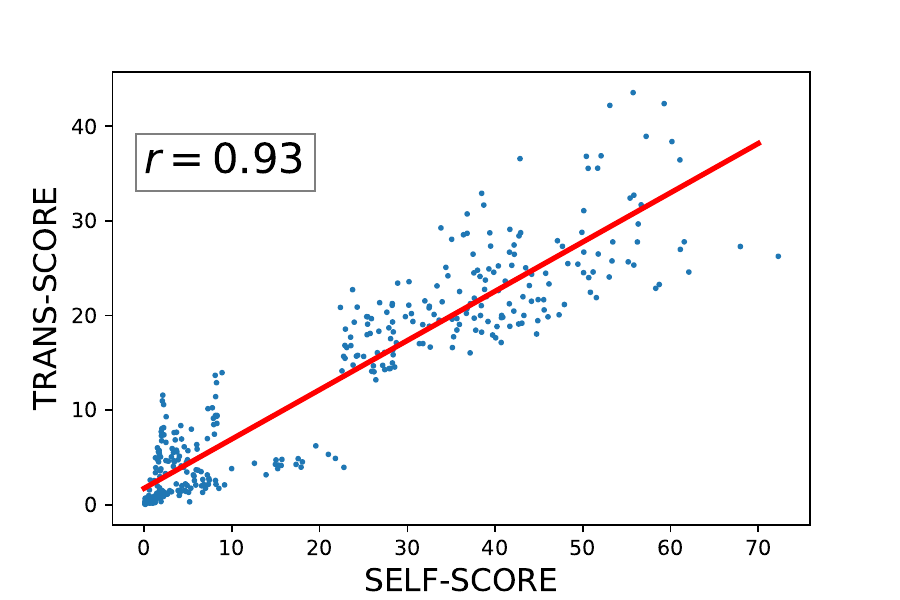}} 
&\subcaptionbox{\label{c}}{\includegraphics[width=0.2\linewidth]{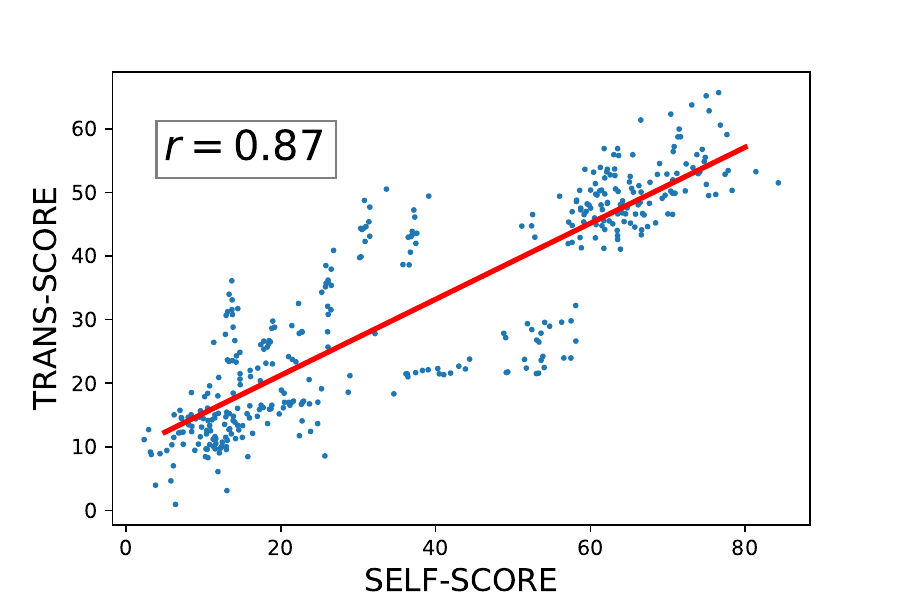}} &
\subcaptionbox{\label{d}}{\includegraphics[width=0.2\linewidth]{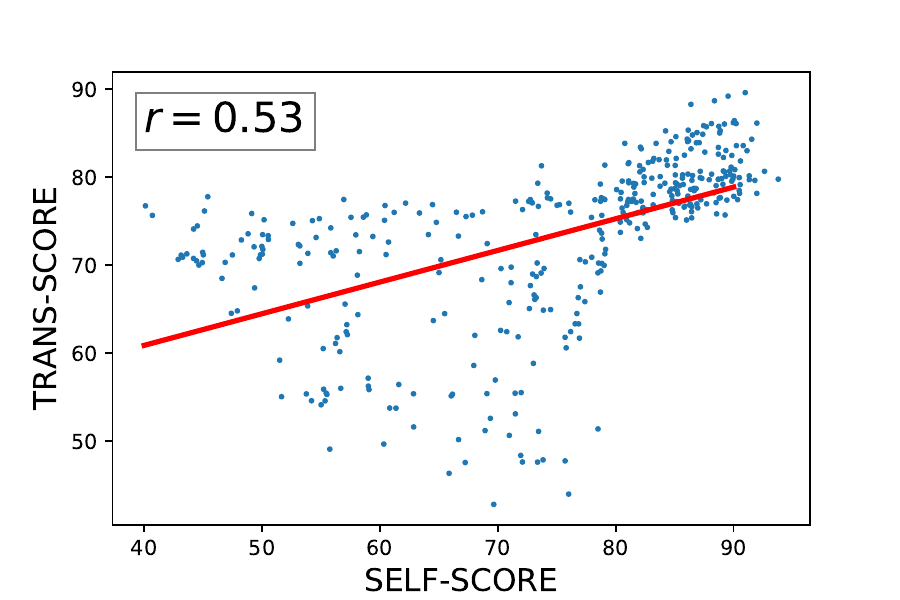}}
\\

\subcaptionbox{\label{a}}{\includegraphics[width=0.2\linewidth]{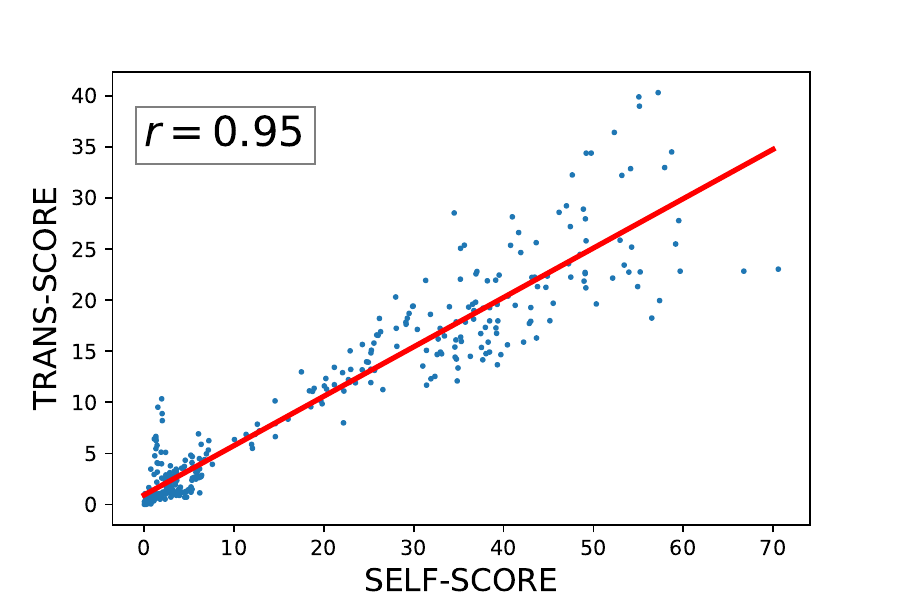}} &
\subcaptionbox{\label{b}}{\includegraphics[width=0.2\linewidth]{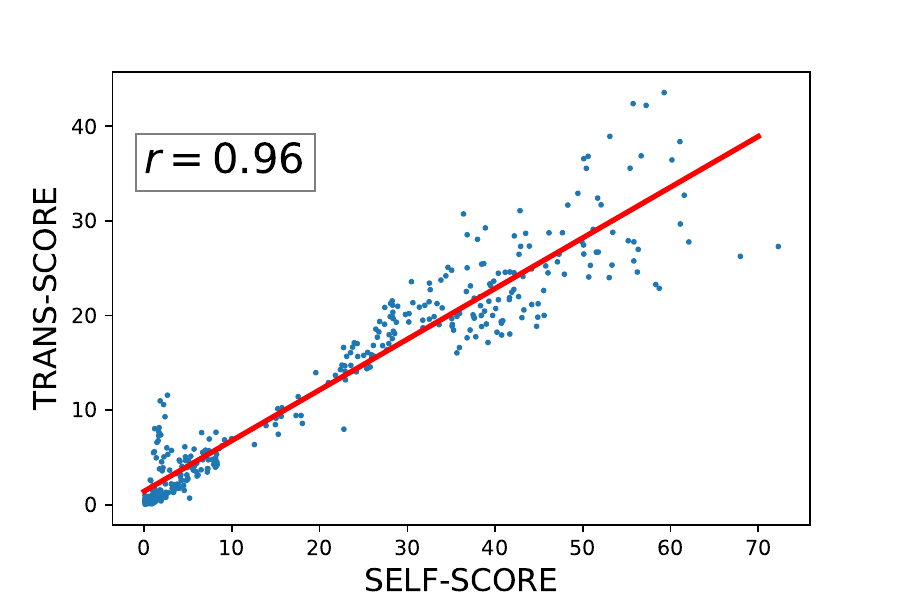}} &
\subcaptionbox{\label{c}}{\includegraphics[width=0.2\linewidth]{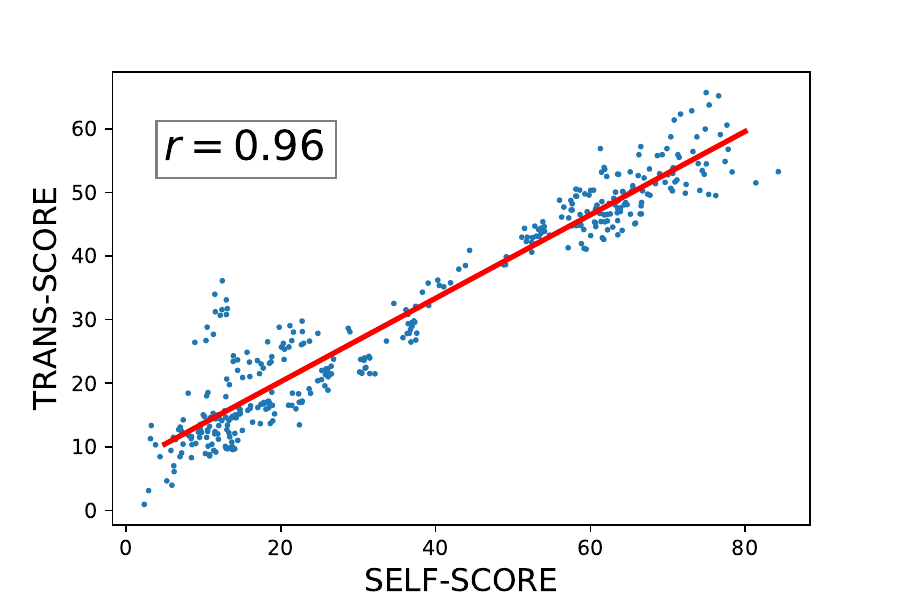}} &
\subcaptionbox{\label{d}}{\includegraphics[width=0.2\linewidth]{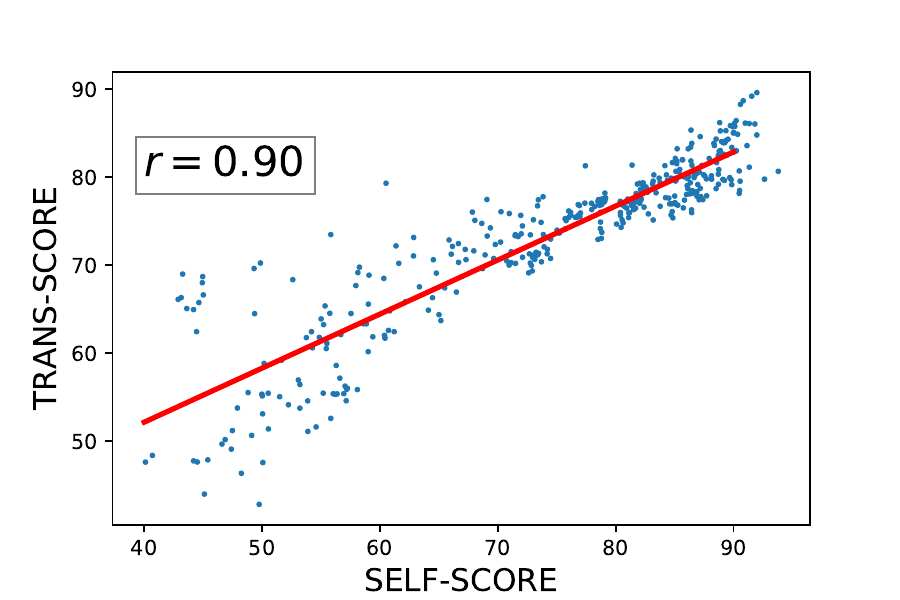}}\\
\end{tabular}
\caption{\protect\raggedright The first row is the correlations between $\text{\selfscore}_{A\circlearrowright B}^{\mathcal{M}}$ and $\text{\transscore}_{A\rightarrow B}^{\mathcal{M}}$ on \mmbase using (a) \sacrebleu, (b) \spbleu;, (c) \chrf and (d) \bertscore. The second row is the correlations between $\text{\selfscore}_{B\circlearrowright A}^{\mathcal{M}}$ and $\text{\transscore}_{A\rightarrow B}^{\mathcal{M}}$ on \mmbase using (e) \sacrebleu, (f) \spbleu, (g) \chrf and (h) \bertscore. All experiments with overall Pearson's $r$.}
\label{fig:mmbase}
\end{figure*}

%% file: tables/pearson_r_correlations.tex
\begin{table*}[!ht]
\centering
\small
\begin{tabular}{lc|cccc}
\hline 
MT System  & Comparison & \sacrebleu & \spbleu & \chrf & \bertscore \\ \hline
\multirow{2}{*}{\mbart} & $A\rightarrow B$ vs. ${A \circlearrowright B}$ &0.78&0.86&0.63&0.53\\
& $A\rightarrow B$ vs. ${B \circlearrowright A}$ &0.94 &0.94 &0.96 & 0.88\\
\hline
\multirow{2}{*}{\mmbase} & $A\rightarrow B$ vs. ${A \circlearrowright B}$ &0.83&0.93&0.87&0.53 \\
& $A\rightarrow B$ vs. ${B \circlearrowright A}$ &0.95 &0.96 & 0.96&0.90\\
\hline
\end{tabular}
\caption{Pearson's $r$ between $\transscore_{A \rightarrow B}^{\mathcal{M}}$ and $\selfscore^{\mathcal{M}}$ (both $A \circlearrowright B$ and $B \circlearrowright A$) using different automatic evaluation metrics $\mathcal M$ on \typeI test set of \dataset.}
\label{tab:correlation_all}

\end{table*}



%% file: tables/ablation.tex
\begin{table*}[!ht]
\centering
\resizebox{1.0\textwidth}{!}{
\begin{tabular}{ll|ccc|ccc|ccc}
\hline
\multirow{2}{*}{\textbf{MT System}} 
& \multirow{2}{*}{\textbf{Self-Trans Feature}} 
& \multicolumn{3}{c|}{\textbf{Type I}} 
& \multicolumn{3}{c|}{\textbf{Type II}} 
& \multicolumn{3}{c}{\textbf{Type III}} 

\\ \cline{3-11} 
& & \textbf{MAE} $\downarrow$ & \textbf{RMSE} $\downarrow$ & 
\textbf{$r$} $\uparrow$
& \textbf{MAE} $\downarrow$ & \textbf{RMSE} $\downarrow$ & 
\textbf{$r$} $\uparrow$ 
& \textbf{MAE} $\downarrow$ & \textbf{RMSE} $\downarrow$ & 
\textbf{$r$} $\uparrow$ 

\\ \hline
\multirow{4}{*}{\mbart}
& \spbleu (basic model) & 2.13 &2.99 &0.94 & 1.61 & 2.19 & 0.93 & 1.20 & 1.38 & 0.94\\
& \ \ + \MaxCount & 2.01 & 2.92 & 0.94 & 1.54 & 2.15 & 0.94 & 1.12 & 1.34 & 0.94\\
& \ \ + \RefLen & 2.07 & 2.96 & 0.94 & 1.61 &2.21 & 0.93 & 1.17 & 1.45 & 0.94\\
& \ \ + \MaxCount \& \RefLen & 2.00 &2.92 & 0.94 & 1.53 & 2.16 & 0.94 & 1.08 &1.33 & 0.95\\
\cline{1-11}
\multirow{4}{*}{\mmbase}
& \spbleu (basic model) & 3.97 &5.72 &0.96 & 3.24 & 4.18 & 0.96 & 3.18 & 3.88 & 0.95\\
& \ \ + \MaxCount & 2.95 & 4.00 & 0.96 & 2.74 & 3.67 & 0.95 & 2.82 & 3.62 & 0.93\\
& \ \ + \RefLen & 3.61 & 5.32 & 0.96 & 2.93 &3.92 & 0.96 & 2.90 & 3.67 & 0.94\\
& \ \ + \MaxCount \& \RefLen & 2.95 & 4.10 & 0.96 & 2.71 & 3.65 & 0.95 & 2.79 & 3.59 & 0.93\\
\hline
\end{tabular}
}
\caption{The results of using auxiliary features to \spbleu for training predictors. We test the performance of \mbart and \mmbase cross language pairs in \typeI, \typeII and \typeIII of \dataset.}
\label{tab:rq_5}
\end{table*}

%% file: tables/synthetic_results.tex
\begin{table}[!h]
\centering
\resizebox{1.0\linewidth}{!}{
\begin{tabular}{ccccc}
\hline
\textbf{Langauge Pair}
& \textbf{MAE} $\downarrow$ & \textbf{RMSE} $\downarrow$ & \textbf{K.} $\boldsymbol \tau$ $\uparrow$ & \textbf{P.} $\boldsymbol r$ $\uparrow$\\
\hline
de-fr &2.21 & 2.67&1.00 & 0.98 \\
en-ta &0.88 & 0.98 &1.00 & 0.99 \\
zh-en & 1.69& 2.37&1.00 & 0.99 \\
\hline
Average &1.59  & 2.01&1.00 & 0.99\\
\hline
\end{tabular}
}
\caption{Results of prediction and ranking on translation quality of \wmtnews synthetic data for three language pairs.}
\label{tab:rq4_synthetic_data}
\centering
\end{table}

%% file: tables/wmt20_bio.tex
\begin{table}[!h]
\centering
\resizebox{1.0\linewidth}{!}{
\begin{tabular}{c|cccc|cccc}
\hline
\multirow{2}{*}{\textbf{Langauge Pair}} 
& \multicolumn{4}{c|}{$B \circlearrowright A$} &\multicolumn{4}{c}{$A \circlearrowright B$ \& $ B \circlearrowright A$}\\
\cline{2-9}
& \textbf{MAE} $\downarrow$ & \textbf{RMSE} $\downarrow$ & \textbf{K.} $\boldsymbol \tau$ $\uparrow$& \textbf{P.}  $\boldsymbol r$ $\uparrow$
& \textbf{MAE} $\downarrow$ & \textbf{RMSE} $\downarrow$ & $\textbf{K.} \boldsymbol \tau$ $\uparrow$& \textbf{P.}  $\boldsymbol r$ $\uparrow$\\
\hline
de-en &10.96 &11.06 &0.80 &0.75 &10.15 &10.21 &0.80 &0.76\\
en-de & 5.41 &5.69  &0.80 &0.63 &5.94 &6.06&0.80&0.63\\
en-es &6.42&7.95  &0.80 &0.82 &6.31 &7.42 &0.80 &0.83\\
en-fr &4.03 &6.27 & 0.40 &0.19 &3.68 &5.86 &0.40 &0.20\\
en-it &6.13 &6.92 &0.40 &0.56 &5.94 &6.58 &0.40 &0.57\\
en-ru & 4.16 &5.62 &0.20 &0.46 &4.20 &5.18 &0.20 &0.49\\
en-zh &2.17 &2.73 &0.20 & -0.04 &2.21 &2.59 &0.00 &0.02 \\
es-en &6.58 &8.17 &0.60 & 0.75 &6.23 &7.48 &0.80 &0.79\\
fr-en &6.12 &8.02 &0.60 & 0.66 &5.77 &7.13 &0.60 &0.67\\
it-en &6.33 &7.94 &0.60 &0.50 &5.90 &7.13 &0.60 &0.56\\
ru-en &5.94 &8.51 &0.40 &0.18 &5.51 &7.81 & 0.20 &0.23\\
zh-en & 5.67 &8.15 & 0.20 & 0.22 &5.18 &7.48 &0.20 &0.23\\
\hline
Average& 5.83&7.25 & 0.50 & 0.47 &5.59 &6.74 &0.48&0.50 \\
\hline
\end{tabular}
}
\caption{Results of our predictors on ranking the selected MT systems on \wmtbio shared tasks.} 
\label{tab:wmt20bio_results}
\centering
\end{table}